\documentclass[a4paper,fleqn]{cas-dc}

\usepackage[authoryear]{natbib}

\usepackage{microtype}
\usepackage{graphicx}
\usepackage{subfigure}
\usepackage{booktabs}
\usepackage{multirow}
\usepackage{fontawesome}
\usepackage{arydshln}

\def\tsc#1{\csdef{#1}{\textsc{\lowercase{#1}}\xspace}}
\tsc{WGM}
\tsc{QE}
\tsc{EP}
\tsc{PMS}
\tsc{BEC}
\tsc{DE}

\begin{document}
\let\WriteBookmarks\relax
\def\floatpagepagefraction{1}
\def\textpagefraction{.001}
\shorttitle{Event-based Optical Flow: ANN \textit{vs.} SNN Comparison}
\shortauthors{Y.Xu et~al.}

\title [mode = title]{Event-based Optical Flow on Neuromorphic Processor: ANN \textit{vs.} SNN Comparison based on Activation Sparsification}


\author[1]{Yingfu Xu}[orcid=0000-0001-7834-3204]

\cormark[1]
\ead{yingfu.xu.94@gmail.com}
\credit{Conceptualization, Methodology, Software, Validation, Formal analysis, Investigation, Data Curation, Writing - Original Draft, Visualization}

\author[2]{Guangzhi Tang}[orcid=0000-0002-0204-9225]
\ead{guangzhi.tang@maastrichtuniversity.nl}
\credit{Conceptualization, Methodology, Writing-review \& editing}

\author[3]{Amirreza Yousefzadeh}[orcid=0000-0002-2967-5090]
\ead{a.yousefzadeh@utwente.nl}
\credit{Conceptualization, Methodology, Formal analysis, Investigation, Funding acquisition, Supervision, Writing – review \& editing}

\author[4]{Guido C. H. E. {de Croon}}[orcid=0000-0001-8265-1496]
\ead{g.c.h.e.decroon@tudelft.nl}

\credit{Conceptualization, Methodology, Supervision, Writing - Review \& Editing}

\author[1]{Manolis Sifalakis}[orcid=0000-0002-0949-2094]
\ead{manolis.sifalakis@imec.nl}

\credit{Conceptualization, Methodology, Formal analysis, Writing - Review \& Editing, Supervision, Project administration, Funding acquisition}


\affiliation[1]{organization={Hardware-Efficient Artificial Intelligence Team, Stichting imec Nederland},
                addressline={High Tech Campus 31}, 
                city={Eindhoven},
                postcode={5656 AA}, 
                country={the Netherlands}}

\affiliation[2]{organization={Department of Advanced Computing Sciences, Maastricht University},
                addressline={Paul-Henri Spaaklaan 1}, 
                city={Maastricht},
                postcode={6229 EN}, 
                country={the Netherlands}}

\affiliation[3]{organization={Faculty of EEMCS, University of Twente},
                addressline={Drienerlolaan 5}, 
                city={Enschede},
                postcode={7522 NB}, 
                country={the Netherlands}}

\affiliation[4]{organization={Faculty of Aerospace Engineering, Delft University of Technology},
                addressline={Kluyverweg 1}, 
                city={Delft},
                postcode={2629HS}, 
                country={the Netherlands}}
                






\cortext[cor1]{Corresponding author}



\begin{abstract}
Spiking neural networks (SNNs) for event-based optical flow are claimed to be computationally more efficient than their artificial neural networks (ANNs) counterparts, but a fair comparison is missing in the literature.
In this work, we propose an event-based optical flow solution based on activation sparsification and a neuromorphic processor, SENECA. SENECA has an event-driven processing mechanism that can exploit the sparsity in ANN activations and SNN spikes to accelerate the inference of both types of neural networks.
The ANN and the SNN for comparison have similar low activation/spike density ($\sim$5\%) thanks to our novel sparsification-aware training. 
In the hardware-in-loop experiments designed to deduce the average time and energy consumption, the SNN consumes 44.9ms and 927.0$\rm \mu$J, which are 62.5\% and 75.2\% of the ANN's consumption, respectively. We find that SNN's higher efficiency attributes to its lower pixel-wise spike density (43.5\% \textit{vs.} 66.5\%) that requires fewer memory access operations for neuron states.

\end{abstract}






\maketitle

\section{Introduction}

Estimating optical flow using an event camera is robust to motion blur and varying illumination thanks to the event stream that captures pixel brightness changes asynchronously in high dynamic ranges.
Similar to most computer vision tasks, more and more accurate event-based optical flow estimation has been achieved by deep neural networks \cite{zhu2018ev, gallego2019focus, zhu2019unsupervised, gehrig2021raft, shiba2022secrets, Paredes-Valles_2023_ICCV}, referred to as artificial neural networks (ANNs) in this paper. Spiking neural networks (SNNs) have different operating mechanisms from ANNs that are closer to the biological neural circuits. Instead of using scalar activations to communicate between neurons, neurons of an SNN maintain their membrane voltages (neuron states) across time steps and communicate by binary activations (spikes).
An event camera is known as a neuromorphic sensor given its sparse, asynchronous, and spike-form measurements. Therefore, SNN potentially fits the event camera better. Researchers have been working on using SNNs for event-based optical flow \cite{lee2020spike, chaney2021self, hagenaars2021self, kosta2023adaptive, cuadrado2023optical, ponghiran2023event, schnider2023neuromorphic}. Although the gap is small, SNNs predict less accurate optical flow than ANNs. 
On the other hand, it is widely believed that SNNs have advantageous computational efficiency. 

Because spikes are binary, SNNs require only accumulate (AC) operations while ANNs require multiply-and-accumulate (MAC) operations for a synaptic operation, \textit{i.e.} the integration of an input activation into the neuron \cite{lemaire2022synaptic}. 
In addition, SNN spikes have a sparse nature.
Neuromorphic processors \cite{furber2020spinnaker, davies2021advancing, tang2023seneca, modha2023neural} exploit the sparsity by skipping unnecessary synaptic operations caused by zero inputs to neurons.
Based on the two facts above, works (especially algorithmic ones) \cite{sengupta2019going, deng2020rethinking} compared the number of synaptic operations of ANN and SNN to support the claim that SNNs are computationally more efficient. Many works of SNNs for event-based optical flow \cite{lee2020spike, chaney2021self, hagenaars2021self, kosta2023adaptive, ponghiran2023event} took the measured data from \cite{horowitz20141} that MAC is 5.1$\times$ more expensive than AC to calculate energy cost. 

However, only considering the difference between the computational costs of AC and MAC is not convincing. More comprehensive ANN \textit{vs.} SNN comparison works keep emerging but
there are three major ignored facts in existing comparisons, regarding the task, processor, and activation sparsity of ANNs.
First, the tasks in ANN \textit{vs.} SNN comparisons \cite{sengupta2019going, lee2021accurate, lemaire2022synaptic} were often image classification, where converting the scalar pixel intensity into a series of spikes is required and thus SNNs need multiple forward-passes to produce a prediction while ANNs need only one. Besides, the SNNs were often converted from a trained ANN. There has been no regression task discussed yet. 
Second, either only ANNs were benchmarked on an accelerator \cite{dampfhoffer2022snns} or the processors were different for SNNs and ANNs \cite{lemaire2022synaptic}, harming the fairness of comparison. 
Third, although zero activations of ANNs were considered in energy calculation \cite{lee2021accurate, dampfhoffer2022snns}, there is no work yet comparing SNNs with sparsified ANNs, even if there are fruitful research results in increasing \cite{georgiadis2019accelerating, kurtz2020inducing, grimaldi2023accelerating, li2023lazy, runwal2023parameter} and exploiting \cite{cao2019seernet, kurtz2020inducing, yang2021dynamic, grimaldi2023accelerating} ANN activation sparsity.

In this work, we aim to address the above problems by proposing an efficient neuromorphic solution for event-based optical flow estimation,
which is a representative regression task, filling in the gaps in existing works. It has wide application fields including power-sensitive mobile robots \cite{paredes2023fully} and always-on sensing devices, where energy-efficient neuromorphic processors are preferred. 
Unlike a frame-based camera filming pixel intensity, an event camera allows SNNs to directly use its spike-form measurements. Therefore, both ANNs and SNNs need only one forward-pass for a prediction.
To address the issue of different processors for ANN and SNN, we adopt a neuromorphic processor, SENECA \cite{yousefzadeh2022seneca, tang2023seneca}, that supports both network types with the same processing logic, enhancing the fairness of the comparison. 
As for the ANN's activation sparsity, we sparsify both ANN activations and SNN spikes before comparison. To do so, we propose an effective activation sparsification approach requiring only one training attempt.
Experimental data of performance, \textit{i.e.} network prediction accuracy and cost of time and energy, measured onboard the neuromorphic processor are analyzed to find out how activation/spike density and spatial distribution affect networks' performance. 
This is the first work showing an SNN's advantageous energy and time efficiency over an ANN in a regression task of event-based vision by a fair comparison supported by experimental measurements on a neuromorphic processor.
In the rest of this paper, we use \textit{ac./sp.} to refer to ANN activation and SNN spike when a statement applies to both of them, for simplicity. The term \textit{ac./sp.} density indicates the percentage of non-zero elements in the output tensors of layers. 
The main contributions are summarized as follows:

\begin{itemize}
\item We propose using an activation function with a channel-level thresholding mechanism and surrogate gradience to train the thresholds. This approach significantly reduces \textit{ac./sp.} density for both ANN and SNN to $<$5\%, without harming event-based optical flow prediction accuracy. It is efficient regarding training efforts, requiring only single-stage training and one additional hyperparameter.
\item To pursue a fair comparison, an ANN and an SNN with very similar architectures, numbers of parameters, and \textit{ac./sp.} density are implemented onboard the same multi-core neuromorphic processor, SENECA \cite{yousefzadeh2022seneca, tang2023seneca}. 
\item We discover the different spatial distributions of ANN activations and SNN spikes. The fact that the SNN produces fewer pixels with spikes results in fewer memory accesses and thus higher time and energy efficiency.
\end{itemize}

\section{Related Works}

\subsection{ANN \textit{vs.} SNN Comparison} \label{subsec:related_comparison}

\begin{table*}[] 
\setlength\tabcolsep{1.75pt}
    \caption{Related works on ANN \textit{vs.} SNN comparison. }
    \label{table:related_comparison}
    \begin{center}
    \begin{tabular}{ccccccc}
        \toprule
        \multirow{2}{*}{Work} & Multi. & SNN  & Memory & ANN & \multirow{2}{*}{Processor(s)} & Maximum \\
         & Pass & Training & Energy & Optimization & & Spikes \\
        \midrule

        \cite{deng2020rethinking} & \faCheck \&\faTimes & Direct Train \cite{wu2019direct} & \faTimes & - & - & -, - \\ [0.2cm]

        \multirow{2}{*}{\cite{davidson2021comparison}} & \multirow{2}{*}{\faCheck} & \multirow{2}{*}{-} & \multirow{2}{*}{\faCheck} & \multirow{2}{*}{-} & SpiNNaker & \multirow{2}{*}{1.72, -} \\
         & & & & & \cite{rhodes2018spynnaker} & \\ [0.2cm]

        \multirow{2}{*}{\cite{lemaire2022synaptic}} & \multirow{2}{*}{\faCheck} & \multirow{2}{*}{ANN-SNN Conversion} & \multirow{2}{*}{\faTimes} & \multirow{2}{*}{-} & 2 for ANN, & \multirow{2}{*}{4.0, \faCheck \&\faTimes} \\
         & &  & & & 2 for SNN. &  \\ [0.2cm]
         
        \multirow{2}{*}{\cite{lee2021accurate}} & \multirow{2}{*}{\faCheck} & \multirow{2}{*}{ANN-SNN Conversion} & \multirow{2}{*}{\faCheck} & \multirow{2}{*}{Acti. Spars.} & SCNN & \multirow{2}{*}{-, \faTimes} \\
         & & & & & \cite{parashar2017scnn} & \\ [0.2cm]
         
        \multirow{2}{*}{\cite{dampfhoffer2022snns}} & \multirow{2}{*}{\faCheck} & \multirow{2}{*}{ANN-SNN Conversion} & \multirow{2}{*}{\faCheck} & Acti. Spars. & Eyeriss v2 & \multirow{2}{*}{0.37, \faTimes} \\
         & & & & \& Data Reuse & \cite{chen2019eyeriss} & \\ [0.2cm]
         
        \multirow{2}{*}{ours} & \multirow{2}{*}{\faTimes} & Direct Train \cite{wu2019direct}, & \multirow{2}{*}{\faCheck} & Acti. Spars.* & SENECA & \multirow{2}{*}{-, \faCheck} \\
         & & \cite{hagenaars2021self} & & \& Data Reuse & \cite{xu2024optimizing} & \\
        \bottomrule
    \end{tabular}
    \end{center}
\end{table*}

Some key characteristics of ANN \textit{vs.} SNN comparison works are shown in Table \ref{table:related_comparison}. From left to right, the items are: 
whether multiple forward-passes were required for an SNN prediction; 
how the SNN(s) was trained; 
whether the energy cost of memory access (reading weights, reading and writing neuron states) was considered; 
what techniques were adopted to optimize ANN inference; 
what processor was adopted for neural network inference; 
the maximum average number of spikes received per synapse for the SNN(s) to be more energy efficient than its ANN counterpart(s). 
The meaning of other items in the table are introduced as follows: 
``\faCheck \&\faTimes'' means the studied SNNs have different results; 
``-'' means the item was not clearly stated; 
``\faCheck'' or ``\faTimes'' following the maximum spikes data indicates whether the studied SNNs are more energy-efficient; 
``ANN-SNN Conversion'' means the SNN(s) was converted from trained ANN(s);
``Acti. Spars.'' indicates that the activation sparsity of ANNs was exploited to reduce energy cost. When an activation is zero, all computations and memory accesses needed for the synaptic operation can be saved. The ``*'' on our work means that we not only exploit but also increase the \textit{ac./sp.} sparsity.

It was shown in \cite{dampfhoffer2022snns} that memory access is much more expensive than computation (>95\% of total energy cost), thus using AC instead of MAC has a small impact on the overall energy cost.
About the processors in \cite{lemaire2022synaptic}, 4 accelerators for sequential and paralleled processing are synthesized on FPGA. However, the SNN accelerator does not support parallel processing for convolution layers while the ANN accelerator does, harming the fairness of comparison. 
The highlight of \cite{deng2020rethinking} is that it is the only work that studied tasks of event cameras. It was found that SNNs are less accurate and require more computation when the inputs are images and SNNs require multiple forward-passes. On the contrary, when the inputs are from event cameras and SNNs do a single forward-pass, SNNs are better in both accuracy and computation cost. The shortcoming is that only the number of operations was shown.
Similarly, our work studied an event-based vision task where an SNN achieves better efficiency but, compared with \cite{deng2020rethinking}, we show more comprehensive experimental results based on accelerating the inference of both ANN and SNN.

\subsection{Activation Sparsification} \label{subsec:related_sparsification}
Exploiting and increasing ANN activation sparsity is mainly motivated by improving the network inference speed on CPUs.
Authors of \cite{yang2021dynamic} proposed to rank the channels of a feature map based on their activation magnitudes. Only channels with top rankings will participate in the computations of the following layers. 
Similarly, for the outputs of multi-layer perceptrons, all other than the largest $k$ (hyperparameter) entries are set to zero in \cite{li2023lazy}.
It was found that inducing activation sparsity improved training robustness to label noise and image noise, and led to better prediction confidence. 
Instead of sorting activations by magnitudes, the authors of \cite{cao2019seernet} 
proposed to skip the computations that are predicted to produce zero activations by a quantized version of the same network layer. 
Note that the works mentioned above \cite{yang2021dynamic, cao2019seernet, li2023lazy} induced computational overhead to exploit activation sparsity. To avoid the overhead, our solution directly ignores activations smaller than the trainable activation thresholds that stay constant during inference.

As for increasing ANN activation sparsity,
the training scheme proposed by \cite{grimaldi2023accelerating} produces a set of individual pixels that are forced to have zero activations on all of their channels, leading to columns of the activation matrix skipped when using the im2col-based general matrix multiply (GEMM) technique and thus accelerates computation. 
A $L$1 loss that regularizes the activation map was adopted to punish big activation values and thus produce more zero and near-zero activations \cite{georgiadis2019accelerating}. 
After fine-tuning with this regularizer, activation density was decreased to $\sim$50\% of the original model and the accuracy slightly increased. 
Similarly, authors of \cite{runwal2023parameter} used the hyperbolic tangent (TanH) function with a scaling parameter to sparsify the activation maps of rectified linear unit (ReLU) and a differentiable approximation of the $L$0 norm for other activation functions.
In \cite{kurtz2020inducing}, Hoyer regularization \cite{hoyer2004non} was applied to activations of a pre-trained network during fine-tuning.
Besides, the forced activation threshold ReLU activation function (FATReLU) (Eq. \eqref{eq:fatrelu_ours}) was proposed to replace ReLU. 
Notably, FATReLU is not differentiable. Therefore, \cite{kurtz2020inducing} performed a manual binary search of desired FATReLU thresholds based on whether the network accuracy can be recovered by retraining after applying the thresholds.
Significant GPU hours and human labor are required by the manual search, especially for deep networks with many layers. 
To avoid such big workloads, we propose an approach (Subsection \ref{subsec:trainableThreshold}) that makes the FATReLU thresholds trainable, requiring only a single training attempt.

\section{Networks and Hardware Implementation}

\subsection{FireNet Architecture}

\begin{figure*}[!b] 
\begin{center}
    \centerline{
    \includegraphics[scale=0.58]{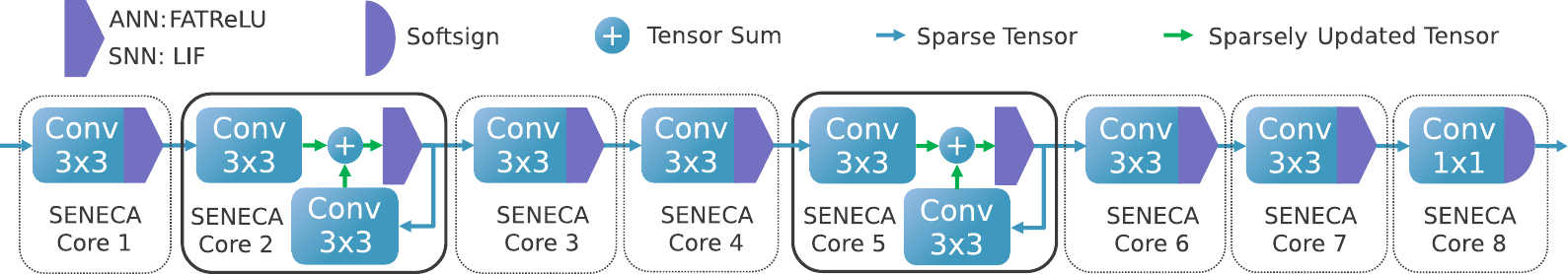}}
    \caption{FireNet network architecture used in this work for event-based optical flow prediction. As stated at the top left of this figure, for ANN, the output tensors of \textit{conv} layers are input to the FATReLU activation function. For SNN, the spike tensors are integrated into the membrane potential of the LIF neurons as synaptic input currents. 
    When the input tensor of a \textit{conv} layer is sparse (blue arrow), a considerable number of pixel locations of the output tensor are not updated by the input tensor. A green arrow indicate such a sparsely updated tensor.
    In \cite{hagenaars2021self}, the flow prediction layer has the TanH activation function. We replace it with Softsign due to its more efficient implementation on SENECA. 
    }
\label{fig:net_arch}
\end{center}
\end{figure*}

The network architecture we study is shown in Fig. \ref{fig:net_arch}. It is based on the FireNet in \cite{hagenaars2021self}. The ANN version and the SNN version have the same architecture of six convolution (\textit{conv}) layers and two recurrent convolution (\textit{rnn-conv}) blocks. 
All \textit{conv} layers are single-strided and have 3-by-3 kernels except the last layer for flow prediction, which has 1-by-1 kernels. All intermediate tensors have 32 channels.

For ANN, a \textit{conv} layer has non-zero trainable biases.
To have a fairer comparison with the SNN, we made two adaptations from the RNN-FireNet (with vanilla ConvRNN) in \cite{hagenaars2021self}. Firstly, we use FATReLU \cite{kurtz2020inducing} activation function (to be introduced in Subsection \ref{subsec:trainableThreshold}) instead of ReLU to pursue higher activation sparsity. Secondly, we modified the vanilla ConvRNN block used by \cite{hagenaars2021self}, shown in Appendix \ref{sec:Old_Recurrent_Block}. As shown in Fig. \ref{fig:net_arch}, our \textit{rnn-conv} block (deployed in SENECA Core 2 and 5) has two \textit{conv} layers instead of three, and the TanH activation function is replaced by FATReLU. This modification reduces the model size and results in sparse information instead of dense one remembered within the \textit{rnn-conv} block. The computational demands for the recurrent \textit{conv} layer (shown below the Tensor Sum icon) and the following layer are greatly reduced thanks to the sparse hidden state. Moreover, as to be shown in Subsection \ref{subsec:accuracy}, our smaller-sized and sparser FireNet models have higher accuracy than the original versions reported in \cite{hagenaars2021self}.
The SNN FireNet in this work is the same as the LIF-FireNet in \cite{hagenaars2021self}, all its \textit{conv} layers have zero biases. 

The presentation of the event camera measurements consists of per-pixel and per-polarity event counts, same as \cite{hagenaars2021self}. It gets populated with consecutive, non-overlapping partitions of the event stream. For training, each input partition (referred to as an event frame) contains a fixed number of events (1,000). For testing, events within a certain time duration are accumulated in an event frame.

\subsection{Processing Mechanism of SENECA} \label{subsec:SENECA}

SENECA \cite{tang2023seneca, yousefzadeh2022seneca} is a programmable digital neuromorphic processor designed with a scalable number of cores. 
Each SENECA core consists of a high-bandwidth SRAM data memory. In this work, we use two sizes for data memory, 256 Kilobytes and 2 Megabytes, for different image resolutions, 56$\times$56 pixels and 120$\times$120 pixels, respectively. The resolution is constrained by the size of data memory due to the need to store the neuron states.
A SENECA core has 8 neuron processing elements (NPEs) that operate in a vector-like fashion. The NPEs are hardware functional units that are time-multiplexed to perform neuron activity computations, \textit{e.g.} addition, multiplication, comparison of numerical values, read from memory, write to memory, \textit{etc}. A SENECA core has a RISC-V controller to conduct other operations, \textit{e.g.} decoding received \textit{event}\footnote{In the context of communication between SENECA cores, an \textit{event} refers to the inter-core message. An \textit{event} has 32 bits, In this work, the directions of \textit{event} flows are shown in Fig. \ref{fig:net_arch}. To distinguish, we refer to the measurements of event cameras as \textit{camera spikes}.}, calculating the addresses of the neuron states and weights, encoding \textit{ac./sp.} into events, \textit{etc}. More details can be found in Appendix \ref{sec:event_seneca}, \ref{sec:cnn_seneca} and \cite{tang2023seneca, xu2024optimizing}.

We map a \textit{conv} layer or a \textit{rnn-conv} block to a SENECA core, as shown in Fig. \ref{fig:net_arch}. Eight SENECA cores are cascadedly connected. 
In the processing of the first \textit{conv} layer, the input \textit{camera spikes} are pre-sorted in spatial order. \textit{Camera spikes} that are closer to the upper edge of the frame are received earlier. For \textit{camera spikes} on the same row, the ones closer to the left are received earlier.
Once \textit{camera spike(s)} at a pixel is received, the neuron states within the patch of pixels whose receptive fields cover the \textit{camera spike}'s pixel location are located and updated accordingly. \textit{Conv} layers of FireNet have 3-by-3 kernels, so a spike updates $3\times 3 \times 32$ neurons in a 3-by-3 pixel patch. When the newly coming spike is on a different pixel, the program switches to the new set of neuron states according to the new pixel location.
Because spikes come in spatial order, it is deterministic whether there will be possible future spikes falling in the receptive field of neurons at a pixel location. 
If not, it means that a pixel location is fully updated and thus the neuron states at this location can be input to FATReLU (for ANN) or compared with the firing threshold (for SNN).
SNN spikes and ANN activations bigger than FATReLU thresholds will then be sent to the following layer in another SENECA core by SENECA \textit{events}. Smaller ANN activations are ignored. Thus SENECA exploits ANN's activation sparsity.
Since the neurons are fired in spatial order, the \textit{ac./sp.} sent to the following layer are in the same spatial order. The following layer in another core starts processing as soon as receiving the first \textit{ac./sp.}. At the same time, the processing of the first layer is still ongoing, leading to parallelization of the layers/cores that requires less network inference time cost than single-thread processing.
This processing mechanism is called event-driven depth-first convolution. It is illustrated in detail in Appendix \ref{sec:cnn_seneca}. 

The \textit{rnn-conv} block's two \textit{conv} layers are processed one after another. The processing of the recurrent layer (shown below the Tensor Sum icon in Fig. \ref{fig:net_arch}) starts after finishing the forward layer on the primary data pass (shown on the left the Tensor Sum icon in Fig. \ref{fig:net_arch}). It is possible to map the recurrent layer in another core to parallel the two \textit{conv} layers of the \textit{rnn-conv} block. The latency (required time to get all optical flow predictions) remains the same but it could reduce the time cost of the \textit{rnn-conv} block to finish all processing. We leave it to future works. 
To reduce the number of memory-accessing operations, it is natural to perform data reuse. SENECA reuses neuron states by a mechanism called \textit{ac./sp.} grouping.
\textit{Ac./sp.} at the same pixel location can be processed together because, in convolutional operation, they update the same set of neurons. We set the group size to four. Firstly, the neuron states are loaded from the memory. Then, four \textit{ac./sp.} are integrated into the neuron states. Lastly, the neuron states are stored back in memory. In this way, the number of memory access operations required to process four events is reduced from four to one. 
If there are fewer than four \textit{ac./sp.} at a pixel location, dummy zero \textit{ac./sp.} fill(s) the group.
In addition to neuron states, an \textit{ac./sp.} is used by all 8 NPEs to multiply with 8 weight parameters and then update 8 neurons in 8 channels of a pixel location. Weights are not reused by our processing mechanism.

\section{Activation Sparsification Methodology} \label{sec:sparsificationMethodologies}

In this section, we describe the approaches to obtain higher \textit{ac./sp.} sparsity for both types of networks. Note that Subsection \ref{subsec:trainableThreshold} and \ref{subsec:FATReLUThresholdInit} apply to ANN. Subsection \ref{subsec:sparsificationRegularizers} applies to both ANN and SNN.

\subsection{Trainable Thresholds for Activations (ANN)} \label{subsec:trainableThreshold}

This subsection introduces the activation function of the ANN.
A ReLU activation function sets all negative activations to zeros and reserves positive activations. As a variant of ReLU, forced activation threshold ReLU (FATReLU) proposed in \cite{kurtz2020inducing} further sets positive activations that are smaller than the threshold $T$ to zeros. 
By adjusting the threshold values, it is possible to reach higher activation sparsities while preserving the network prediction accuracy. 
FATReLU's mathematical expression is 
\begin{equation} \label{eq:fatrelu_ours}
\begin{aligned}
        FATReLU_T(x) &= x \cdot BS_T(x), \\
        BS_T(x) &= \left\{
        \begin{array}{cc}
             1 \quad & if \quad x > T \quad (T>0), \\ 
             0 \quad & otherwise.
        \end{array} 
        \right.
\end{aligned}
\end{equation}
$FATReLU_T(x)$ is discontinuous and not differentiable when $x=T$. So, unlike ReLU, a network with FATReLU cannot be directly trained. 
To avoid the big workload of manual search of $T$ \cite{kurtz2020inducing} as introduced in Subsection \ref{subsec:related_sparsification}, we borrow the idea of surrogate gradient from SNN training.
As shown by Eq. \eqref{eq:fatrelu_ours}, FATReLU is the multiplication of the input activation $x$ and a binary step function $BS_T$ with threshold $T$.
$BS_T$ is the same as the spiking function of SNN. We select the derivative of the inverse tangent as the surrogate gradient function for $BS_T$, same as \cite{hagenaars2021self}. In this way, $T$ becomes a trainable network parameter. We do channel-wise FATReLU thresholding, which means that there are in total $7\times32$ FATReLU threshold parameters in FireNet. All thresholds are initialized to a small value (1e-6) before training.

\subsection{Sparsification Regularizers (ANN and SNN)} \label{subsec:sparsificationRegularizers}

From Eq. \eqref{eq:fatrelu_ours}, we can notice that smaller FATReLU inputs $x$ and bigger FATReLU thresholds $T$ can increase the number of zeros. Similarly, smaller neuron membranes and bigger firing thresholds can lead to fewer spikes of SNN. Therefore, for both ANN and SNN, we add two sparsification regularizers to punish big neuron membranes and encourage thresholds $T$ to grow, respectively. We use $L$1 loss for neuron membrane and $L$2 loss for $T$.
The total training loss is
\begin{equation} \label{eq:sparse_loss}
\begin{aligned}
    L &= L_{flow} + \lambda_{s} \cdot L_{s}, \\
    L_{s} &=  \sum_{i} \lambda_{i} \cdot \left (  \sum \mathrm{ReLU} \left (\mathbf {x}_i \right ) + \sum_{j} \left (\frac{1}{T_{i,j}} \right )^2 \right ),
\end{aligned}
\end{equation}
where $L_{s}$ is the loss made of the two activation sparsification regularizers.
$i$ is the layer index, and $j$ is the channel index. $\mathbf {x}_i$ is the neuron membrane tensor of layer $i$. $\sum \mathrm{ReLU} (\mathbf {x}_i)$ is the summation of all the elements of $\mathrm{ReLU}(\mathbf {x}_i)$. $T_{i,j}$ is the FATReLU threshold of ANN or the firing threshold of a LIF neuron.

\subsection{Threshold Initialization based on Prior Activations (ANN)} \label{subsec:FATReLUThresholdInit}

Using the surrogate gradient for FATReLU thresholds and loss function shown in Eq. \eqref{eq:sparse_loss}, an activation-sparse ANN can be trained from scratch by a single training attempt. In practice, we find that the initialization of FATReLU thresholds affects the network performance. Thus, we propose initializing the FATReLU thresholds based on the activation statistic of a trained network. Specifically, all thresholds have fixed zero values in the first training attempt, and the network is trained from scratch by $L_{flow} + \sum_{i} \lambda_{i} \cdot ( \sum \mathrm{ReLU}  (\mathbf {x}_i ))$. Then, we run the trained network (inference only) on a subset of the training set to log the activation tensors. Based on the logged tensors, the initial threshold $T_{N,M,init.}$ for the $M$th channel of the $N$th layer is set as the median of all logged activations at the $M$th channel of the $N$th layer, \textit{i.e.}, the network is initialized to have 50\% non-zero activations of the original trained network, statistically. The second training attempt fine-tunes the trained network with the initialized thresholds by Eq. \eqref{eq:sparse_loss}. 

\section{Evaluation of Activation Sparsification} \label{sec:eval_acti_spar}

\subsection{Network Training} \label{subsec:networkTraining}

We train our optical flow FireNet based on the open-sourced code of \cite{hagenaars2021self}. The regularizers for activation sparsification are added to the event deblurring loss for self-supervised learning of optical flow $L_{flow}$, as shown in Eq. \eqref{eq:sparse_loss}. A reader who is interested in $L_{flow}$ and the training strategies can refer to \cite{hagenaars2021self}.
The network is trained for 100 epochs on the UZH-FPV dataset \cite{delmerico2019we}. We use the AdamW optimizer with a weight decay of 0.01 and the OneCycleLR scheduler with a maximum learning rate of 2e-4.
In our practice, networks are trained with different $\lambda_{s}$. 
It ranges from 1.0e-6 to 2.6e-6 for ANN and 1.6e-7 to 2.6e-7 for SNN. As for $\lambda_{i}$, the layer with the highest \textit{ac./sp.} density has $\lambda_{i}$=2.0 while other layers have $\lambda_{i}$=1.0. The purpose is to reduce the difference in densities between layers to avoid one or several layer(s) becoming much denser than others and becoming bottleneck(s) of inference speed.

\subsection{Network Accuracy and Activation Density} \label{subsec:accuracy}

\begin{table*}[!b]
\setlength\tabcolsep{1.8pt}
    \begin{center}
    \caption{Accuracy indicated by average endpoint error (AEE) and \textit{ac./sp.} density (Dens.(\%)) of the networks tested on the MVSEC \cite{zhu2018multivehicle} dataset (dt=4). $\%_{\rm Out.}$ is the percentage of outlier optical flow predictions. ``-S'' means that the network is trained with the activation sparsification loss (Eq. \eqref{eq:sparse_loss}). ``-S*'' means that the network has FATReLU with trainable thresholds but it is trained with the loss $L_{flow} + \sum_{i} \lambda_{i} \cdot ( \sum \mathrm{ReLU}  (\mathbf {x}_i ))$, without the regularizer for bigger FATReLU thresholds. ``-FT'' means that the network is trained by fine-tuning a trained model as described in Subsection \ref{subsec:FATReLUThresholdInit}. Other networks without ``-FT'' are trained from scratch. 
    For all metrics, lower is better. The best of each metric is marked by \textbf{bold} text.}
    \label{table:accuracy_and_density}
    \begin{tabular}{cccccccccccc}
        \toprule
        \multirow{2}{*}{Network}
        & \multicolumn{2}{c}{outdoor\_day1}  & \multicolumn{2}{c}{indoor\_flying1} & \multicolumn{2}{c}{indoor\_flying2} & \multicolumn{2}{c}{indoor\_flying3} & \multicolumn{3}{c}{Average} \\
        \cmidrule(r){2-3} \cmidrule(r){4-5} \cmidrule(r){6-7} \cmidrule(r){8-9} \cmidrule(r){10-12}
        & AEE & $\%_{\rm Out.}$ & AEE & $\%_{\rm Out.}$ & AEE & $\%_{\rm Out.}$& AEE & $\%_{\rm Out.}$& AEE & $\%_{\rm Out.}$& Dens.($\%$) \\
        \midrule
        EV-FlowNet(GRU)\cite{hagenaars2021self} & \textbf{1.69} & 12.50 & 2.16 & 21.51 & 3.90 & 40.72 & 3.00 & 29.60 & 2.94 & 29.35 & - \\ 
        \hdashline
        RNN-EV-FlowNet & \textbf{1.69} & 12.96 & \textbf{2.02} & \textbf{18.74} & 3.84 & 38.17 & 2.97 & \textbf{27.91} & 2.88 & \textbf{27.32} & 16.90 \\ 
        RNN-EV-FlowNet-S (smaller $\lambda_{s}$) & 1.92 & 17.34 & 2.06 & 18.83 & \textbf{3.56} & \textbf{37.02} & \textbf{2.88} & 28.94 & \textbf{2.79} & 27.76 & 5.35 \\ 
        RNN-EV-FlowNet-S (bigger $\lambda_{s}$) & 1.73 & \textbf{12.20} & 2.03 & 19.03 & 3.83 & 39.70 & 3.02 & 30.58 & 2.90 & 28.71 & 4.78 \\ 
        \hdashline
        LIF-EV-FlowNet & 1.99 & 15.99 & 2.47 & 26.79 & 4.94 & 50.51 & 3.91 & 39.59 & 3.68 & 37.47 & 9.46 \\ 
        LIF-EV-FlowNet-S (smaller $\lambda_{s}$) & 2.01 & 16.88 & 2.69 & 32.00 & 4.77 & 51.29 & 3.84 & 41.85 & 3.66 & 39.90 & 5.81 \\ 
        LIF-EV-FlowNet-S (bigger $\lambda_{s}$) & 1.88 & 16.36 & 2.76 & 33.63 & 4.96 & 52.65 & 4.06 & 44.76 & 3.80 & 41.68 & \textbf{3.93} \\ 
        \midrule
        
        FireNet(GRU)\cite{hagenaars2021self} & 2.04 & 20.93 & 3.35 & 42.50 & 5.71 & 61.03 & 4.68 & 53.42 & 4.41 & 49.92 & - \\ 
        \hdashline
        RNN-FireNet & 1.94 & 17.80 & 3.11 & 38.79 & 5.45 & 57.31 & 4.47 & 49.59 & 4.19 & 46.22 & 34.03 \\ 
        RNN-FireNet-S* (no threshold regularizer) & \textbf{1.67} & \textbf{12.88} & \textbf{2.79} & \textbf{32.70} & \textbf{5.02} & \textbf{51.99} & \textbf{4.05} & \textbf{43.69} & \textbf{3.80} & \textbf{40.54} & 16.57 \\ 
        RNN-FireNet-S & 2.16 & 22.04 & 3.16 & 40.09 & 5.14 & 55.96 & 4.24 & 48.76 & 4.05 & 46.25 & 5.92 \\ 
        RNN-FireNet-S-FT & 1.97 & 18.31 & 3.24 & 39.23 & 5.48 & 57.00 & 4.45 & 49.02 & 4.22 & 46.09 & \textbf{4.52} \\ 
        
        \hdashline
        LIF-FireNet & 1.96 & 15.82 & 3.32 & 41.37 & 5.99 & 62.24 & 4.98 & 54.63 & 4.58 & 49.94 & 19.54 \\ 
        LIF-FireNet-S & 2.15 & 21.06 & 3.14 & 39.17 & 5.59 & 57.94 & 4.63 & 51.19 & 4.31 & 47.36 & 4.53 \\ 
        \bottomrule
    \end{tabular}
    \end{center}
\end{table*}

In this work, we focus on the lightweight FireNet architecture. But to show that the proposed activation sparsification approach can generalize, we also applied it to EV-FlowNet, using the code from \cite{hagenaars2021self}. Note that, we do not intend to deploy EV-FlowNet on the neuromorphic processor due to its large size. So we do not modify the architecture of its recurrent block. 
In Table \ref{table:accuracy_and_density}, we took two networks, EV-FlowNet (GRU) and FireNet (GRU), from \cite{hagenaars2021self} as the baselines of network prediction accuracy. The other networks in the table are trained by us. RNN-FireNet has 74,722 parameters. RNN-FireNet-S and RNN-FireNet-S-FT have 74,946 parameters. LIF-FireNet and LIF-FireNet-S has 74,818 parameters. RNN-EV-FlowNet has 23,531,752 parameters. RNN-EV-FlowNet-S has 23,535,240 parameters. LIF-EV-FlowNet and LIF-EV-FlowNet-S have 20,400,840 parameters.
Surprisingly, our networks with vanilla RNN have better accuracy than the baselines with GRU. We attribute it to our training scheme with the AdamW optimizer and the OneCycleLR learning rate scheduler.

\textit{Ac./sp.} density is calculated for the output \textit{ac./sp.} tensors of the input layer and hidden layers. The \textit{ac./sp.} density of each layer is logged for all testing samples. The average \textit{ac./sp.} density of a network model is the average over all its layers and all testing samples. The output layer has a dense activation function, SoftSign for our adapted FireNet and Tanh for EV-FlowNet. The RNN blocks of EV-FlowNet have a dense Tanh activation function. We do not apply activation sparsification to such layers and do not take the 100\% activation density into account when calculating the average activation density. The density of input \textit{camera spikes} is not taken into account either, since it depends on the dataset and thus is not correlated to the network models. 

In Table \ref{table:accuracy_and_density}, we show two models of the sparsified ANN EV-FlowNet with vanilla RNN (RNN-EV-FlowNet-S) and the sparsified SNN LIF-EV-FlowNet (LIF-EV-FlowNet-S), respectively. ``Smaller $\lambda_{s}$'' and ``bigger $\lambda_{s}$'' mark two separate training attempts of the same network with the same \textit{ac./sp.} sparsification approach but different weights for sparsification loss ($\lambda_{s}$). 
The accuracy and \textit{ac./sp.} density of the two network models tell that \textit{ac./sp.} sparsification with smaller $\lambda_{s}$ produces similar accuracy but higher \textit{ac./sp.} sparsity than the network model without sparsification. \textit{Ac./sp.} sparsification with bigger $\lambda_{s}$ can train network models with slightly lower accuracy but higher \textit{ac./sp.} sparsity. Accuracy-sparsity trade-off is discussed in the next subsection.


As shown in the bottom half of Table \ref{table:accuracy_and_density}, RNN-FireNet-S* has the highest accuracy. It is close to the accuracy of LIF-EV-FlowNet-S which has 20,400,840 parameters, $\sim$272.2 times of RNN-FireNet-S (74,946 parameters). The activation density of RNN-FireNet-S* is around half of RNN-FireNet. It indicates that lower density could exist together with higher accuracy. This phenomenon was also observed by works \cite{georgiadis2019accelerating, kurtz2020inducing, yang2021dynamic, hoefler2021sparsity, runwal2023parameter}. We intuitively explain this phenomenon by that the sparsification regularizers can suppress the noise in activations. Only the activations carrying important information are kept. The ``denoised'' sparse activations can contribute to accuracy. 
RNN-FireNet-S-FT has only $\sim$13.3\% activations of RNN-FireNet while maintaining a similar accuracy. For LIF-FireNet-S, we observe a small improvement in accuracy and 23.2\% density of the LIF-FireNet trained without sparsification. 
In Section \ref{sec:hardware_experiments}, RNN-FireNet-S-FT and LIF-FireNet-S are compared with hardware in the loop.

\subsection{Ablation Study} \label{subsec:ablation}

To better understand the effects of the sparsification regularizers, we perform an ablation study. 
For simplicity, we use neuron density and pixel density to refer to neuron-wise \textit{ac./sp.} density and pixel-wise \textit{ac./sp.} density, respectively. Pixel density is the percentage of pixel locations that have at least one channel with \textit{ac./sp.}.
For instance, an activation tensor of a \textit{conv} layer has 5$\times$5 pixel locations and 8 channels on each pixel. If there are 16 non-zero elements in the activation tensor, then the neuron-wise activation density is 16/(5$\times$5$\times$8)=8\%. If these 16 non-zero elements are located on 2 pixel locations (all 8 channels are non-zero for these 2 pixels), then the pixel-wise activation density is 2/(5$\times$5)=8\%. If there are 16 pixels each of which has only one non-zero channel, then there are 16 activated pixels and the pixel density is 16/(5$\times$5)=64\%. Therefore, pixel density reflects the spatial density of \textit{ac./sp.} in the 2-dimensional domain of pixel locations, \textit{i.e.} image plane.

\begin{figure*}[!t]
\begin{center}
    \centerline{
    \includegraphics[scale=0.6]{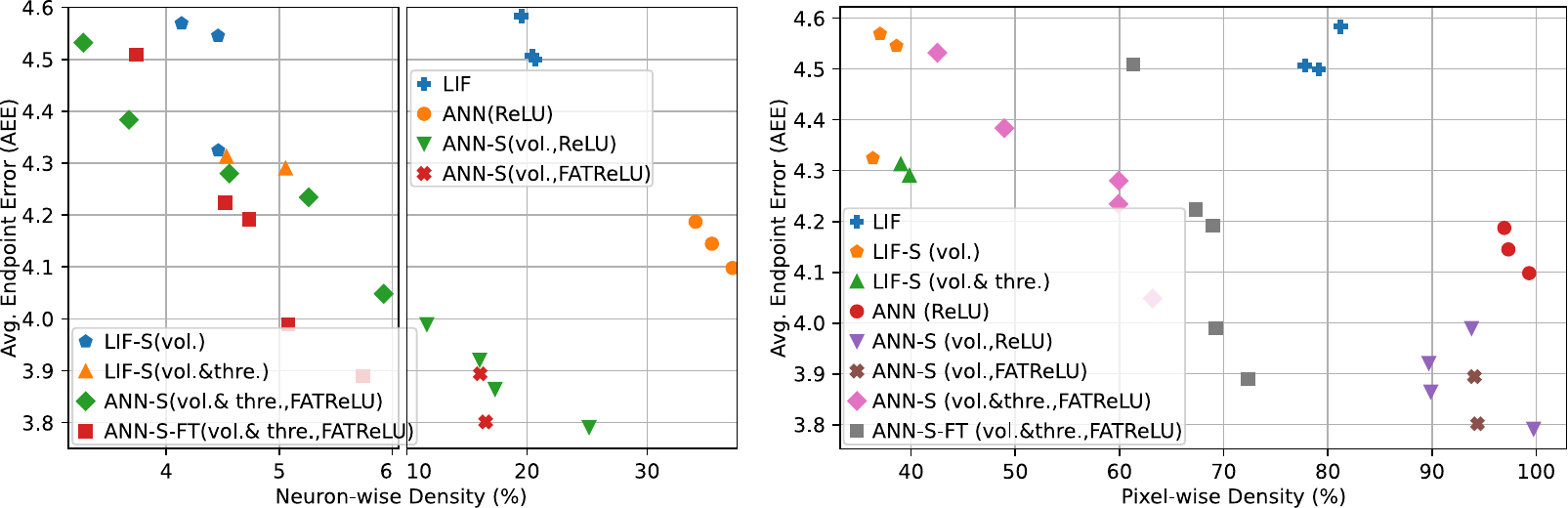}}
    \caption{The correlation between prediction accuracy and \textit{ac./sp.} density. The accuracy metric is the AEE over the whole test set. The shown neuron-wise and pixel-wise density ($x$-axis) is the average of all the layers and all testing samples. The networks shown in both subplots are the same.
    ``\textsl{vol.}'' in the legend means that the networks are trained with the $L$1 regularizer for neuron membrane voltage. ``\textsl{vol.\&thre.}'' means the sparsification loss involves both the $L$1 voltage regularizer and the $L$2 regularizer that encourages the ANN's FATReLU thresholds or SNN's firing thresholds to grow.
    There is no network whose neuron density is between 6\% and 10\% so the $x$-axis of the left subplot is truncated.}
\label{fig:density_aee}
\end{center}
\end{figure*}

The neuron density and average endpoint errors (AEEs) of network models trained with different settings of sparsification loss are shown in the left subplot of Fig. \ref{fig:density_aee}. For simplicity, \textsl{ANN} represents RNN-FireNet and \textsl{LIF} represents LIF-FireNet.
To remove as much of the effect of randomness in network training as possible, we trained $\sim$10 models for each loss setting with different weights on the sparsification loss ($\lambda_{s}$ in Eq. \eqref{eq:sparse_loss}). In the left subplot of Fig. \ref{fig:density_aee}, if the point of \textsl{Model A} is on the lower left side of the point of \textsl{Model B}, it means \textsl{Model A} is better in both accuracy and neuron sparsity. In this case, if the two models are trained with the same loss setting, \textsl{Model B} is omitted in both subplots. 

First, we discuss the information in the left subplot. For ANN, when applying the $L$1 regularizer on neuron membrane voltage alone (\textsl{ANN-S(vol.,ReLU)}, green inverted triangle), the accuracy significantly increases and the neuron density decreases to around 50\% of the networks trained without sparsification loss (\textsl{ANN (ReLU)}, orange circle). 
Comparing \textsl{ANN-S(vol.,FATReLU)} marked by the red crosses and \textsl{ANN-S(vol.,ReLU)}, we can notice that having FATReLU with trainable thresholds brings small improvement to accuracy and the density is little affected. 
After applying the $L$2 regularizer that encourages bigger thresholds, density dramatically drops to $\sim$5\%, as shown by \textsl{ANN-S(vol.\&thre.,FATReLU)} marked by the green diamonds. Comparing the green diamonds with the red squares (\textsl{ANN-S-FT}), we can see that the latter performs better in both metrics, to a small extent.
Similarly, for SNNs with LIF neurons, applying sparsification loss noticeably benefits both density and accuracy. A blue pentagon (\textsl{LIF-S(vol.)}) and an orange triangle (\textsl{LIF-S(vol.\&thre.)}) are closely located, indicating the small effect from the $L$2 loss for the SNN firing thresholds.
Then, we move to the right subplot of Fig. \ref{fig:density_aee} to analyze the pixel density. 
As shown, SNNs have lower pixel densities ($\sim$39\%) than ANNs. ANNs with high accuracy (AEE$<$4.0) have high pixel density ($>$68\%). 

To summarize, \begin{itemize}
\item applying sparsification loss significantly reduces \textit{ac./sp.} density without deteriorating accuracy; 
\item Sparsification can dramatically increase ANNs' accuracy with only $\sim$50\% activations as the ANN without sparsification;
\item For highly sparsified networks whose neuron density is lower than 6\%, a negative correlation exists between accuracy and sparsity;
\item SNN models are generally less accurate than ANNs but have much lower pixel density.
\end{itemize}

\section{Experiments onboard Neuromorphic Processor} \label{sec:hardware_experiments}

We select an ANN (RNN-FireNet-S-FT in Table \ref{table:accuracy_and_density}) and an SNN (LIF-FireNet-S) to deploy onboard SENECA and conduct comparisons. We measure\footnote{
All hardware-related measurements were performed in gate-level simulation using industry-standard ASIC simulation and power measurement tools (Cadence Xcelium and Cadence JOULES) for GF-$22nm$ FDX technology node (in the typical corner $0.8V$ and $25C$, no back-biasing, 500MHz clock frequency). The power results are accurate within 15\% of signoff power and include the total power consumption of the chip, i.e. both dynamic and static power. We have not included the I/O power consumption in the reported results.} the time and energy cost of network inference. We test on several representative input event frames instead of the whole dataset because hardware simulation is very time-consuming. It takes around 30 hours to simulate 70 milliseconds of FireNet inference. 


\subsection{Single Layer Comparisons in Controlled Conditions} \label{subsec:ctrl_experiment}

\begin{figure*}[!t]
\begin{center}
    \centerline{
    \includegraphics[scale=0.65]{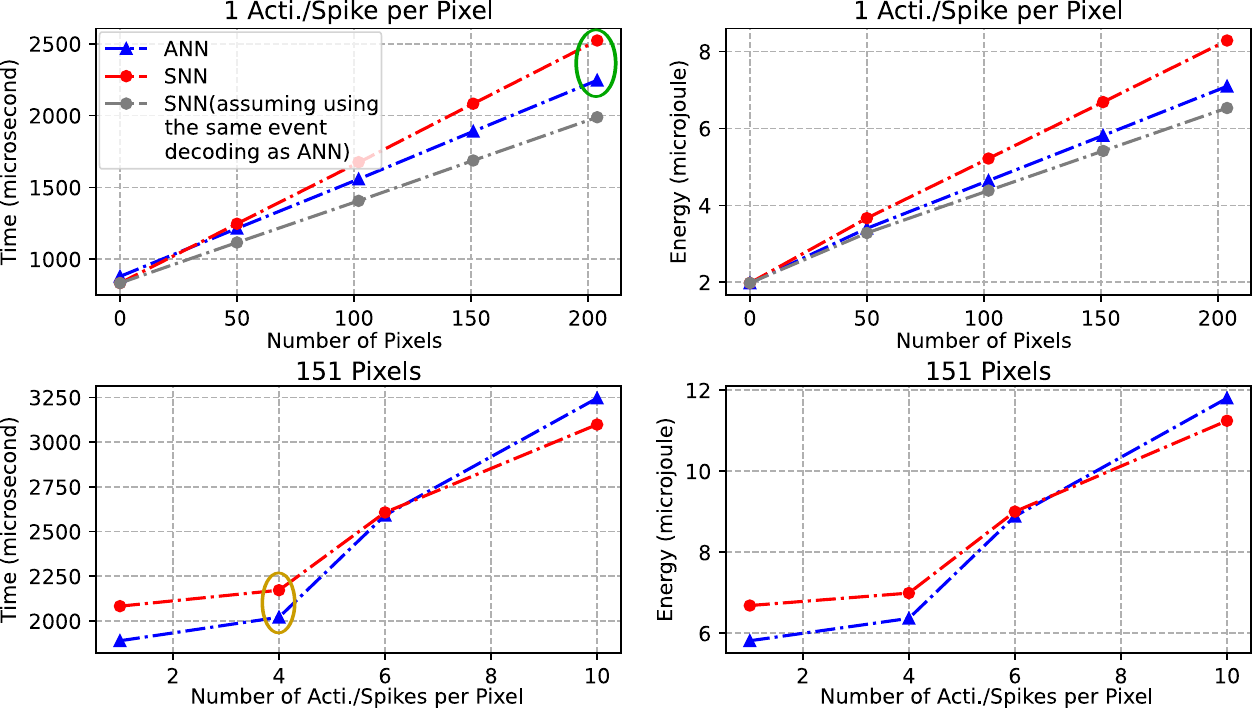}}
    \caption{Time and energy cost of the controlled experiments. In the bottom two subplots, 151 pixels have at least one \textit{ac./sp.}. 
    } 
\label{fig:pixel_density}
\end{center}
\end{figure*}

\begin{figure*}[!t]
\begin{center}
    \centerline{
    \includegraphics[scale=0.72]{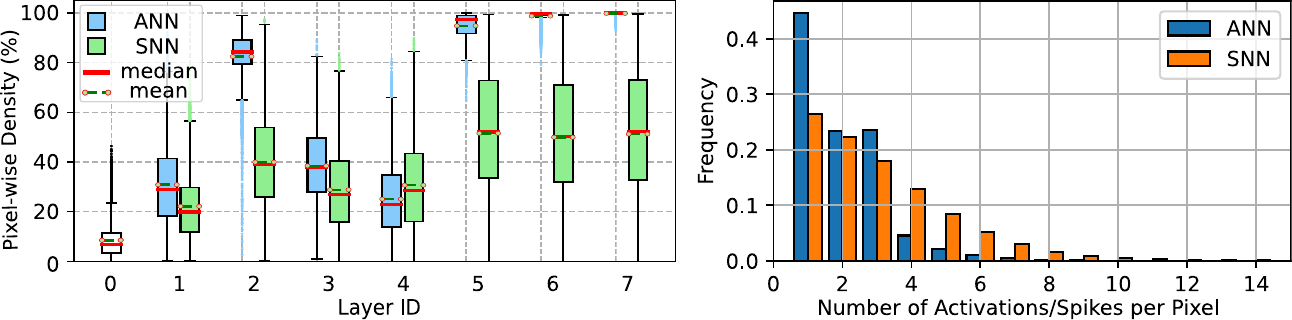}}
    \caption{Layer-wise distribution of pixel density (left) and distribution of the numbers of \textit{ac./sp.} per pixel (right) based on the data logged in testing. 
    }
\label{fig:density_distribution}
\end{center}
\end{figure*}

Before testing the whole network on real data of the test set, we conduct a comparison to study how pixel density and number of \textit{ac./sp.} on the same pixel affect the processing cost of a single \textit{conv} layer. The results are shown in Fig. \ref{fig:pixel_density}. 
In these experiments, the spatial resolution is 16$\times$16, 256 pixels in total. Each pixel has 32 channels, the same as FireNet. We consider the total processing cost of receiving SENECA events, updating neuron states, and comparing the neuron states with the thresholds. Capturing, encoding, and sending the \textit{ac./sp.} are disabled because it is hard to control the number of output \textit{ac./sp.}.
In the first set of experiments (top row of Fig. \ref{fig:pixel_density}), we generate one \textit{ac./sp.} for each pixel because most of the output tensors of FireNet layers have a single \textit{ac./sp.} per pixel, as shown in Fig. \ref{fig:density_distribution}. The numbers of pixels that have an \textit{ac./sp.} are 0, 50, 102, 151, and 204. 
Shown in the top two subplots of Fig. \ref{fig:pixel_density}, there is a clear linear correlation between the number of pixels having \textit{ac./sp.} and the inference time. SNN is more time-consuming than ANN because the decoding schemes of SENECA \textit{events} are different between SNN and ANN. 

For ANN, we use a 32-bit \textit{event} to encode an activation, 16 bits for the channel index (unsigned integer), and the rest 16 bits for the activation value (BFloat16). In contrast, for SNN, we use a bit coding scheme. The 32 bits of an \textit{event} encode the 32 channels. If a channel has a spike, the corresponding bit is 1 otherwise 0. While for ANN, the number of \textit{events} is the same as the number of activations. SNN requires fewer events and thus is more efficient in terms of inter-core communication. 
However, to decode the spikes from the 32-bit \textit{event}, 32 iterations are required to check every bit. When there is one \textit{ac./sp.} at a pixel, ANN takes 2.52$\mu s$ to decode. In contrast, SNN needs 5.14$\mu s$. When the number is 28, SNN takes less time (8.04 \textit{vs.} 8.14). This phenomenon tells us that the software implementation onboard SENECA should take into account the spike distribution of the certain network to select the better \textit{event} encoding scheme. In this work, all experiments of SNN are conducted with the bit coding scheme.
To eliminate the effects of \textit{event} encoding schemes, we subtract the time difference for the SNN assuming it uses the same encoding as ANN, shown by grey circles in the top row of Fig. \ref{fig:pixel_density}.
After the subtraction, the SNN is more efficient. We attribute it to the fact that a binary spike does not require the memory access and multiplication required by an activation.
Note that we use the power measurements of the SNN with the bit encoding scheme when calculating the energy cost. Using ANN's encoding scheme should lead to lower power cost and thus less energy than what the top-right subplot of Fig. \ref{fig:pixel_density} shows.

As shown in the two bottom subplots of Fig. \ref{fig:pixel_density}, increasing the number of \textit{ac./sp.} from 1 to 4 leads to less growth in time and energy cost than increasing it from 4 to 6. It is because of the \textit{ac./sp.} grouping strategy that processes four spikes together, introduced in Subsection \ref{subsec:SENECA}. When the number is below 5, it requires only one neuron state updating operation. When the number increases to 6, it requires two operations thus the cost increases significantly. When the number is 10, three operations are required.
Comparing the data points in the circles in the two subplots on the left, the points in the green circle have 204 \textit{ac./sp.}. The points in the yellow circle have 604. Although having more \textit{ac./sp.}, the time costs of the points in the yellow circle are less, due to the \textit{ac./sp.} grouping mechanism that reduces the required numbers of memory access operations of neuron states. Therefore, higher neuron density does not necessarily mean more time cost, the spatial distribution of \textit{ac./sp.} is an important factor. SENECA processing can be more efficient when the \textit{ac./sp.} are concentrated at fewer pixels.

\begin{figure*}[!hbpt]
\begin{center}
    \centerline{
    \includegraphics[scale=0.248]{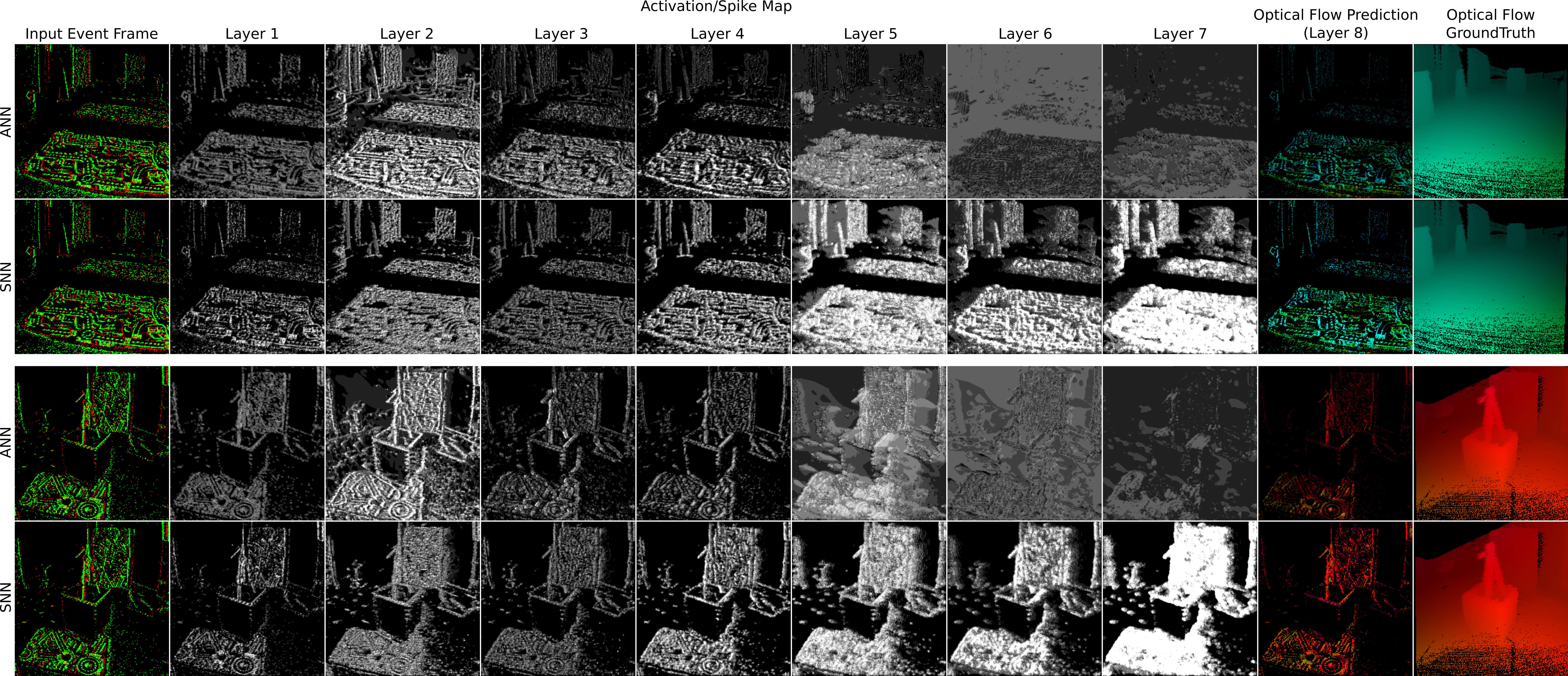}}
    \caption{Visualization of event-based optical flow prediction on two testing event frames. The first column shows the accumulated \textit{camera spikes} (event frames) that are the input to the networks. Events within the time bin of 12.5 milliseconds are accumulated in the image frame. Events in red or green capture brightness changes in two polarities (turn brighter and turn darker). The 2nd to 8th columns show the \textit{ac./sp.} density of each pixel. For layers 1 to 7, their \textit{ac./sp.} tensors have 32 channels. If there are no non-zero \textit{ac./sp.} at a pixel, its color is black. If there are 8 or more than 8 channels that have non-zero \textit{ac./sp.} at a pixel, its color is white. The bigger the number of \textit{ac./sp.}, the brighter the pixel. The second column from the right shows the optical flow prediction from the network. Only pixels that have at least one input event have an optical flow prediction, which is a 2-dimensional vector in the image plane, encoded by colors as shown in Fig. \ref{fig:color_wheel}. The column on the right shows the dense ground truth optical flow measured by other sensors, provided by the testing dataset \cite{zhu2018multivehicle}.}
\label{fig:pixel_dens_vis}
\end{center}
\end{figure*}

\begin{figure}[!hbpt]
\begin{center}
    \centerline{
    \includegraphics[scale=0.4]{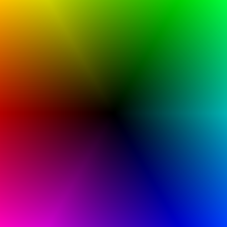}}
    \caption{Optical flow color coding scheme. Direction is encoded in color hue, and speed in color brightness.}
\label{fig:color_wheel}
\end{center}
\end{figure}

\subsection{Spatial Distributions of \textit{Ac./Sp.}}
\label{acti_distributions}

After the experiments of \textit{ac./sp.} distribution of a single layer in controlled conditions, we compare and analyze the \textit{ac./sp.} distributions of the ANN and the SNN on the real test set, and then deduce which network type can be more efficient.
The boxplots reflecting the distribution of pixel density are shown in the left subplot of Fig. \ref{fig:density_distribution}.
ANN has a higher pixel density than SNN in 6 out of 7 layers. For the last 3 layers, almost all inferences have almost 100\% pixel density. 
In the right subplot of Fig. \ref{fig:density_distribution}, we can see that the SNN statistically has more spikes per pixel than the ANN. It is reasonable because the neuron densities of the ANN and SNN are similar and the SNN has lower pixel density. Based on Fig. \ref{fig:density_distribution}, we claim that the SNN is more likely to have higher efficiency because its spikes are concentrated at fewer pixels. As discussed in Subsection \ref{subsec:ctrl_experiment}, this feature benefits processing efficiency.

Fig. \ref{fig:pixel_dens_vis} visualizes the accumulated \textit{camera spikes} (\textit{i.e.} event frame that is the network input), \textit{ac./sp.} maps produced by each layer, and event-based optical flow predictions. The caption for Fig. \ref{fig:pixel_dens_vis} introduced each type of its subfigure in detail. From Fig. \ref{fig:pixel_dens_vis}, we notice that ANN has higher pixel-wise activation density, especially for Layers 5, 6, and 7. Almost every pixel has at least one activation while the number of activation(s) at one pixel is small, as indicated by the dark grey color in the first and third rows of Fig. \ref{fig:pixel_dens_vis} that correspond to the ANN's activation maps. In contrast, Layers 5, 6, and 7 of the SNN produced many pixels without any spike. Most of such dark pixels are in the image regions where there is no input \textit{camera spikes}. For image regions with many \textit{camera spikes}, all layers of the SNN produce dense spikes.
It can be summarized that higher contrast in an \textit{ac./sp.} density map means a higher degree of spatial concentration of \textit{ac./sp.} and higher pixel sparsity. The spike density maps of the SNN have higher contrast than the activation maps of the ANN. Thus SNN has higher pixel sparsity. This observation is consistent with Fig. \ref{fig:density_distribution}.

Fig. \ref{fig:density_distribution} and Fig. \ref{fig:pixel_dens_vis} show the pixel-wise \textit{ac./sp.} density of the FireNet architecture.
One feature of FireNet is that the output tensors of all layers have the same spatial size, \textit{i.e.}, there is no downsampling or upsampling. For EV-FlowNet, a more complicated UNet-like architecture with downsampling encoding and upsampling decoding, we got similar results.
The RNN-EV-FlowNet-S (4th row in Table \ref{table:accuracy_and_density}) has average (over the testing dataset) pixel density (\%) at each layer: [100, 85.39, 100, 100, 99.99, 69.1, 71.65, 65.78, 99.99, 100, 99.98, 100].
In contrast, the layer-wise pixel density (\%) of the LIF-EV-FlowNet-S (7th row in Table \ref{table:accuracy_and_density}) is: [
57.97, 63.69, 79.77, 90.49, 88.09, 70.05, 89.26, 85.07, 89.23, 85.25, 79.54, 67.89]. For 9 out of 12 layers, the ANN (RNN-EV-FlowNet-S) has higher pixel density than its SNN counterpart (LIF-EV-FlowNet-S).

\subsection{Network Experiments} \label{subsec:network_experiments}

Recalling Subsection \ref{subsec:SENECA}, we test two resolutions on two sizes of on-chip memory of SENECA. 
We first show and discuss the time and energy costs of the event frames with 56$\times$56 pixels and then the results of 120$\times$120 pixels. The network prediction accuracy of 120$\times$120 pixels onboard SENECA follows.
For each type of network, we select three event frames from the test set that lead to different neuron densities. For each layer of the SNN, the frames lead to respectively $\sim$20\%, $\sim$100\%, and $\sim$200\% of the average neuron density of the layer over the test set. For the ANN whose neuron density is narrower distributed, the three input frames respectively lead all layers except the 6th layer to have $\sim$50\%, $\sim$100\%, and $\sim$150\% of the average layer-wise neuron density. 
Although the 6th layer does not coincide with expectation, the two selected frames for $\sim$50\% and $\sim$150\% density have the least differences from the expected densities among all frames in the test set.

\begin{figure*}[!t]
\begin{center}
    \centerline{
    \includegraphics[scale=0.68]{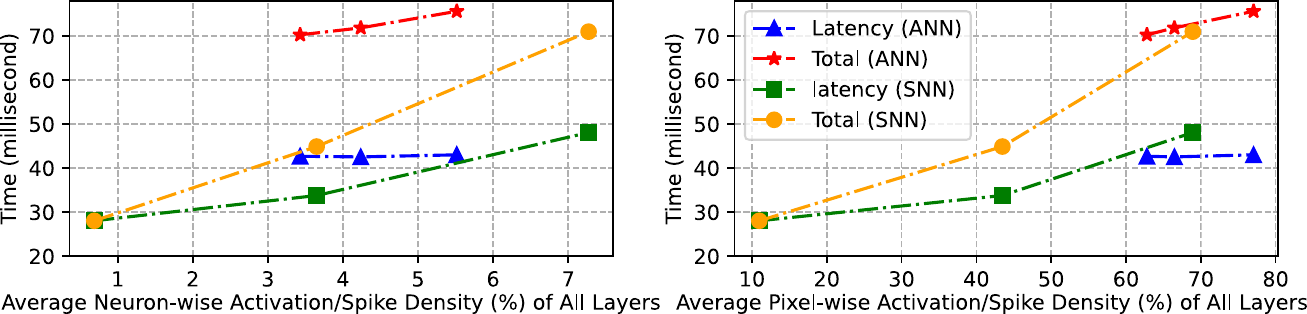}}
    \caption{Correlation between the average \textit{ac./sp.} density of all layers and network inference time cost measured on the three testing frames with 56$\times$56 pixels. The density of the input \textit{camera spikes} is not taken into account for the average density.
    We define the latency of processing a frame as the duration from when the first \textit{camera spike} is input to the network to when all optical flow vectors have been predicted. 
    We define the total time cost as the duration from when the first \textit{camera spike} is input to the network to when all processing including the \textit{rnn-conv} blocks introduced in Subsection \ref{subsec:SENECA} has been finished. The data points in the two subplots are the results of the same testing frame if their horizontal sequence (left-right) is the same.} 
\label{fig:time_dense}
\end{center}
\end{figure*}

Fig. \ref{fig:time_dense} shows the time cost and its correlations to neuron density and pixel density. Same as the results shown in Subsection \ref{subsec:ctrl_experiment}, we see a positive correlation between pixel density and time cost. The three frames of the ANN have similar time costs because ANN's last three layers have almost 100\% pixel density on almost all testing frames (shown in Fig. \ref{fig:density_distribution}), including the selected three frames. These layers took longer time than other layers due to the high pixel density. Because of the parallelization of the layers/cores introduced in Subsection \ref{subsec:SENECA}, those layers became the bottlenecks of the whole network inference. Given the similar pixel densities ($\sim$100\%) of the bottleneck layers of the testing frames, it is reasonable to get similar time costs. 
As shown in the right subplot of Fig. \ref{fig:time_dense}, when the pixel density is $\sim$70\%, the SNN has a larger latency than the testing frames of the ANN. The reason is that the SNN testing frame has noticeably higher neuron density, as shown in the left subplot. 
Recalling the two bottom subplots of Fig. \ref{fig:pixel_density}, more \textit{ac./sp.} on the same pixel leads to longer time.

\begin{table*}[] 
\setlength\tabcolsep{2.0pt}
    \caption{Total energy cost of the Testing Frames (56$\times$56 pixels)}
    \label{table:power}
    \begin{center}
    \begin{tabular}{ccccccc}
        \toprule
        Network and Density & ANN 50\% & ANN 100\% & ANN 150\% & SNN 20\% & SNN 100\% & SNN 200\% \\
        \midrule
        Energy Consum. ($\mu$J)& 1115.0 & 1233.0 & 1340.5 & 391.1 & 927.0 & 1320.0 \\
        \bottomrule
    \end{tabular}
    \end{center}
\end{table*}

\begin{table*}[t] 
\setlength\tabcolsep{2.0pt}
    \caption{Accuracy and \textit{ac./sp.} density (dt=4) of the networks tested on downsampled event frames with 120 $\times$ 120 pixels, running on GPU or SENECA. ``Dens. (\%)'' refers to neuron density. }
    \label{table:seneca_accuracy_120}
    \begin{center}
    \begin{tabular}{ccccccccc}
        \toprule
        \multirow{3}{*}{Sequence}
        &\multicolumn{4}{c}{ANN}& \multicolumn{4}{c}{SNN} \\
        \cmidrule(r){2-5} \cmidrule(r){6-9}
        &\multicolumn{2}{c}{AEE}& \multicolumn{2}{c}{Dens. (\%)}& \multicolumn{2}{c}{AEE}& \multicolumn{2}{c}{Dens. (\%)}\\
        \cmidrule(r){2-3} \cmidrule(r){4-5} \cmidrule(r){6-7} \cmidrule(r){8-9}
        & GPU & SENECA& GPU & SENECA& GPU & SENECA& GPU & SENECA \\
        \midrule
        outdoor\_day1    & 3.23 & 3.08 & 3.65 & 3.62 & 3.82 & 3.75 & 2.04 & 1.99 \\ 
        indoor\_flying1  & 4.18 & 4.05 & 4.22 & 4.17 & 3.68 & 3.65 & 3.62 & 3.52 \\ 
        indoor\_flying2  & 5.73 & 5.67 & 4.67 & 4.62 & 5.60 & 5.61 & 4.88 & 4.78 \\ 
        indoor\_flying3  & 4.77 & 4.71 & 4.50 & 4.45 & 4.67 & 4.67 & 4.41 & 4.30 \\ 
        Average   & 4.77 & 4.68 & 4.40 & 4.35 & 4.62 & 4.61 & 4.13 & 4.03 \\
        \bottomrule
    \end{tabular}
    \end{center}
\end{table*}

Comparing the data points corresponding to the average neuron densities (the one in the middle for each polyline), it is noticeable that the SNN has higher time efficiency. The SNN costs 44.9ms in processing the selected testing frame, which is 62.5\% of 71.8ms, the time cost of the ANN. 
The SNN testing frame for average neuron density leads to pixel density (\%) [9.3, 25.2, 45.3, 30.8, 31.5, 55.4, 56.9, 59.2] for each layer, 9.3 is the density of the input \textit{camera spikes} and the average density of the 7 layers is 43.5\%. It is higher than the average pixel density (\%) over the whole test set, which is [8.6, 22.2, 40.0, 28.9, 30.7, 51.6, 50.0, 51.3], 8.6 is the density of the input \textit{camera spikes} and the average density of the 7 layers is 39.2\%. Therefore, the average time cost of the SNN over the test set could be lower than the selected testing frame.
As for the ANN, the testing frame for average neuron density results in pixel density (\%) [8.8, 28.6, 82.5, 37.4, 22.8, 96.2, 98.3, 99.6] for each layer, 8.8 is the density of the input \textit{camera spikes} and the average density of the 7 layers is 66.5\%. It is slightly lower than the average over the whole test set, [8.6, 30.9, 82.5, 38.5, 25.1, 94.9, 98.8, 99.9], 8.6 is the density of the input \textit{camera spikes} and the average density of the 7 layers is 67.2\%. 
Therefore, the ANN could cost more time than the selected frame, on average over the test set. 
Based on the fact that the SNN's time cost is 62.5\% of the ANN's when processing the selected testing frames for average neuron density, we claim that the SNN is averagely more time-efficient than the ANN on the test set and the average time cost of SNN is less than 62.5\% of the ANN's.
In addition, SNNs can be more time efficient if using the same SENECA \textit{event} encoding scheme as ANN, as discussed in Subsection \ref{subsec:ctrl_experiment}.
The energy costs are shown in Table \ref{table:power}. The SNN's energy cost is 75.2\% of the ANN's.

The event frames with 56$\times$56 pixels are downsampled from the central patch of 112$\times$112 pixels. It only covers 19.14\% area of the original event frame with 256$\times$256 pixels. Besides, we find that, for many frames, there is no available ground truth optical flow for any pixel. Therefore, we only evaluate the network accuracy on event frames of 120$\times$120 pixels on SENECA. For each of the ANN and the SNN, we select a frame producing $\sim$100\% of the average neuron density over the whole test set on all layers. 
The statistical accuracy and density of the whole test set shown in Table \ref{table:seneca_accuracy_120} are obtained by the Python-based SENECA simulator running on a CPU server. This simulator produces exactly the same computation results as SENECA without any precision difference.
SENECA uses 16-bit BFloat16 as the data type, but the network parameters trained on GPU are 32-bit single-precision floating-point numbers. We do post-training quantization that rounds the trained network parameters to BFloat16 values to deploy the networks on SENECA.
We also run the same tests on GPU to compare with the results from SENECA. 
In general, SENECA produces slightly better accuracy and sparsity than GPU, which means that the post-training quantization to BFloat16 does not deteriorate accuracy. Intuitively, we propose the hypothesis that BFloat16 has enough precision in the context of optical flow prediction using FireNet.

The time and energy costs of processing 120$\times$120 pixels are listed as follows. The latency and total time of the ANN are 182.69ms and 326.39ms, respectively. For SNN, the measured durations are 148.91ms and 225.77ms, respectively. The SNN is $\sim$18\% and $\sim$30\% more efficient in latency and total time cost, respectively.
The energy cost of the ANN is 5753.4 $\mu$J and the SNN consumes 4174.4 $\mu$J. The SNN has a $\sim$27\% advantage. 
In general, the advantage of the SNN over the ANN observed in the test on 120$\times$120 pixels is similar to the observation in the test on 56$\times$56 pixels.

\section{Conclusions}
In this work, we proposed an event-based optical flow solution based on \textit{ac./sp.}-sparsified neural networks onboard a neuromorphic processor, and then conduct a fair comparison of an ANN and an SNN with the same lightweight architecture. The major findings follows,
\begin{itemize}
\item \textit{Ac./sp.} sparsification is proven useful for the two network architectures for event-based optical flow (FireNet and EV-FlowNet). Notably, for an ANN trained with activation sparsification, its accuracy can be greatly improved. At the same time, its neuron-wise activation density is greatly reduced;
\item 
Given similar neuron density, lower pixel density means that, firstly, fewer neurons are required to be found (based on the \textit{ac./sp.} pixel location), accessed, and updated. Secondly, the \textit{ac./sp.} grouping mechanism reuses the neuron states to be updated more times because statistically there are more \textit{ac./sp.} on a pixel.
This statement can be generalized to other processing mechanisms that reuse neuron states in a \textit{conv} layer;
\item  The SNN and the ANN in comparison have similar neuron density but the SNN has significantly lower pixel density, which is the main contributor to SNN's higher efficiency. 
It is an interesting topic to study why the ANN has higher pixel density and how to reduce it.
\end{itemize}

This work makes one step forward in the field of SNN \textit{vs.} ANN comparison by studying a more complicated regression task of event-based vision. Although the network we deploy on the neuromorphic processor is more complicated than all the networks in the same type of literature, it is still only a lightweight network without state-of-the-art accuracy.
The obstacle is our current limited capacity of mapping bigger-size networks with more complicated inter-layer connections to the multi-core processor. In the end, we hope this work can encourage the community to find more evidence showing SNN's better efficiency in fair comparisons.

\section*{Appendix}
\appendix

\begin{figure*}[!hbpt]
\begin{center}
    \centerline{
    \includegraphics[scale=0.6]{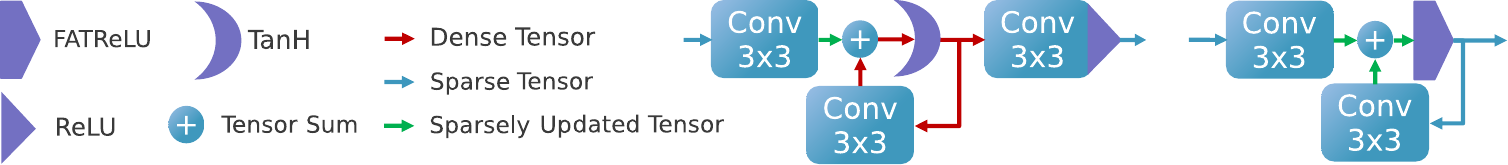}}
    \caption{The vanilla \textsl{ConvRNN} block adopted by \cite{hagenaars2021self} (left) and our recurrent block (right).}
\label{fig:original_ConvRNN}
\end{center}
\end{figure*}

\section{Open-Sourced Code}
The code developed for this work are available at 
\href{https://github.com/YingfuXu/ANN_vs_SNN_evflow }{link}.

\section{ANN Recurrent Block} \label{sec:Old_Recurrent_Block}

The difference between the recurrent block of FireNet we use and the original one \cite{hagenaars2021self} is shown in Fig. \ref{fig:original_ConvRNN}. The original one has one more \textit{conv} layer and uses a TanH activation function. TanH produces a non-zero output for a non-zero input. Thus, the output tensor has almost zero sparsity (red arrow). The main purpose of our modification is to get the ANN and the SNN in comparison to have the same network architecture and sparse feature maps.
In terms of model capacity, our modification is disadvantageous, but our modification greatly increases activation sparsity. In addition, the accuracy of our models with sparsification-aware training does not deteriorate.

\section{Event-Driven Neural Processing on SENECA} \label{sec:event_seneca}

This section was first introduced in \cite{xu2024optimizing}. We include it here and adapt the writing for the context of this work for ease of reading. An interested reader can refer to \cite{xu2024optimizing} for more details.

SENECA is a programmable digital neuromorphic processor that is capable of performing a wide range of tasks. The processor is designed with a scalable number of cores. Each core consists of data memory, a flexible controller (RISC-V), a dedicated controller (loop controller), an event capture unit, multiple (configurable, in this work, 8) neuron processing elements (NPEs) that operate in a vector-like fashion, and a programmable Network on Chip (NoC) which facilitate the event communication among the cores. The NoC delivers a SENECA inter-core events (referred to as \textit{event} for simplicity) to the destination core based on the content of its routing table, which can change dynamically by RISC-V. In this work, given the cascaded layer connection of FireNet, events are sent from one core to another in a single direction.

The NPEs are hardware functional units that are time-multiplexed to perform neuron activity computations.
Each NPE is connected to a high-bandwidth SRAM data memory (16 bits for each NPE) and has a register file (RF) with 64 16-bit words that can be used for computation. This improves energy efficiency as the access energy cost is smaller than the SRAM memory. When in computation mode, all NPEs work in lock-step mode, executing the same instruction at any given cycle similar to a single instruction, multiple data (SIMD) operation.

As a neuromorphic platform, SENECA generates \textit{ac./sp.} outputs through the NPEs when they meet certain conditions according to the workload. In this work, if the neuron state is bigger than the threshold, an \textit{ac./sp.} would be generated.
These \textit{ac./sp.} are then processed by the event capture unit, which converts the input \textit{ac./sp.} vector into address event representation (AER) format \cite{AER2017}. The event capture unit sends an interrupt to the RISC-V controller for further processing whenever a new \textit{ac./sp.} is generated. In this work, generated activation(s)/spike(s) are encoded event(s) and then transmitted to another core through the NoC. 

The RISC-V controller decides which operations should be executed on the NPEs depending on the workload scheduling. The loop controller coordinates the time-multiplexing of NPEs and the address generation for data memory access. It dispatches microcodes to the NPEs, enabling the processing of events. Each microcode is invoked to handle a specific type of event, such as neuron updates, threshold evaluations, or data conversions. For a more in-depth review of the SENECA architecture, please refer to \cite{yousefzadeh2022seneca, tang2023seneca, tang2023open, xu2024optimizing}.

To optimize the processing of sparse data flows between layers of neurons, SENECA executes event-driven neural processing. There are different types of events, where each type triggers a specific set of computations, such as binary spikes produced by spiking neurons, non-zero activations generated by the ReLU/FATReLU activation function, and the synchronization signal representing the end of the time step or data frame. 
When an \textit{ac./sp.} is received, an event-integration task is executed. For ANN, the event-integration task multiplies the activation value with the corresponding weight vector and integrates the results into the neuron state vector. For SNN, the corresponding weight vector is integrated into the neuron state vector. It has been demonstrated in \cite{tang2023seneca} that integrating activation results in minimal energy overhead than integrating binary spikes.
When a synchronization event is received, an event generation task is executed. The neurons that have not been processed by the event generation task yet will be processed. As to be introduced in Section \ref{sec:cnn_seneca}, for a \textit{conv} layer, a neuron would be processed by the event generation task as soon as its state has been updated by the last \textit{ac./sp.} in its receptive field. So a synchronization event will trigger the processing of the rest neurons in the lower part of the feature map. 
For ANN, the event generation task applies the activation function, \textit{e.g.} FATReLU, to neuron states and generates non-zero activation events if there is neuron state(s) passing the thresholding. For SNN, the firing threshold is compared with the neuron membrane voltage to decide whether to generate a binary spike.
    
In general, event-driven processing for neural networks includes three phases:

\begin{itemize}
    \item \textbf{Event Reception}: Unpack/decode the event and prepare for neural processing based on the information carried by the event and the recipient neurons.
    \item \textbf{Neural Processing}: Execute neurosynaptic computations and update neuron states.
    \item \textbf{Event Transmission}: Pack the generated spikes in one event packet and multi-cast it to the destination core(s).
\end{itemize}

During neural network computation, an event received from the NoC wakes up the RISC-V controller in a SENECA core and triggers the event reception phase. According to the decoded event, the event reception function determines the type of neural processing required and defines a set of executable tasks (which are represented by micro-code to be executed at the NPEs). The loop controller receives the tasks and controls the time-multiplexed neural processing steps in the NPEs. The loop controller operates asynchronously with the event reception functions, allowing for accelerated and parallelized event processing. If a task execution involves event generation from neuron states, the event generator collects non-zero outputs from NPEs and packs them as AER events. These events then wake up the RISC-V controller and trigger the event transmission phase, which encodes and packages the AER events as compressed network event packets. Finally, the network event packet is sent to the destination core(s) through the NoC. Neural processing through the loop controller can work in parallel with the event reception/transmission processes since the loop controller can orchestrate the neural processing independently from the RISC-V controller.

\section{Event-Driven Depth-First Convolution on SENECA} \label{sec:cnn_seneca}

Same as the previous section, this section was also introduced in \cite{xu2024optimizing}. 

\begin{figure*}[ht]
\centering
\includegraphics[width=1.0\linewidth]{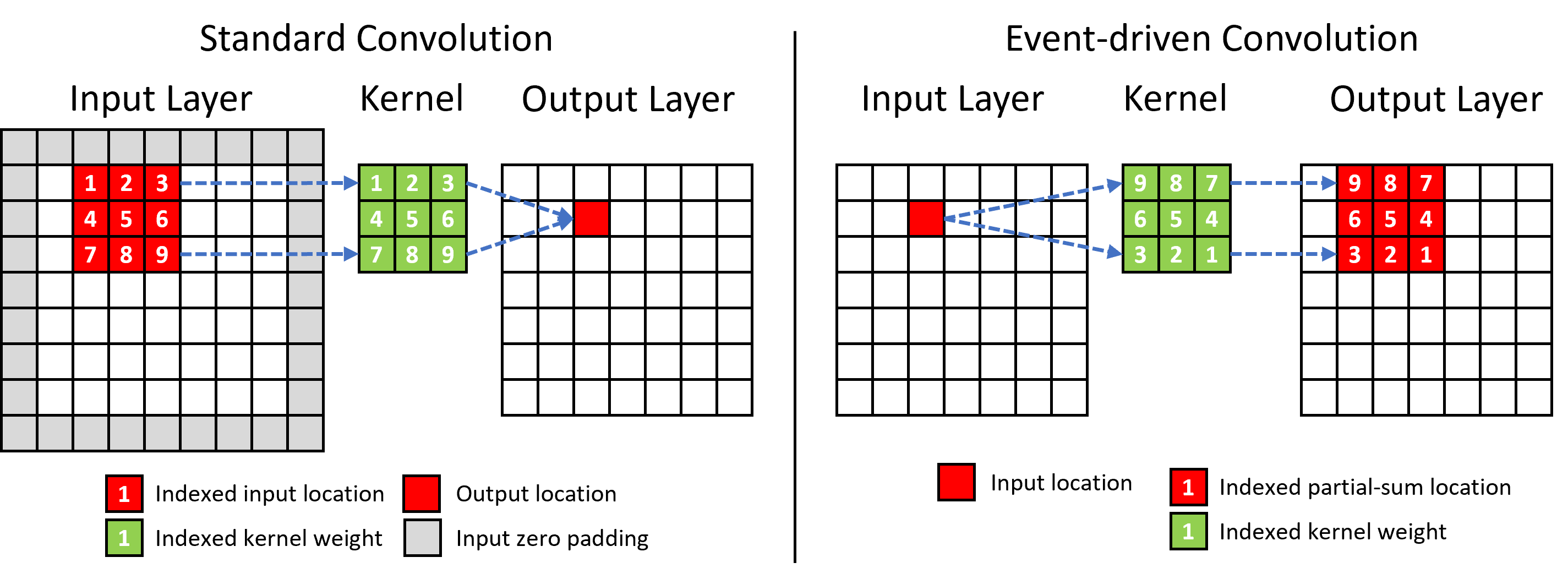}
\caption{Comparison between the standard and the event-driven convolution. The event-driven convolution requires to rearrange the sequence of kernel weights. The change in the spatial sequence for a 3$\times$3 convolution kernel is shown in the figure. The channel dimension of the tensor is omitted for simplicity.}
\label{fig:standard_and_event_conv}
\end{figure*}

Fig. \ref{fig:standard_and_event_conv} shows the differences between the standard and event-driven convolution. The latter processes input \textit{ac./sp.} one by one in their order of arrival and inregrate them incrementally into the neuron states of the corresponding fanned-out postsynaptic neurons. However, this process requires maintaining high-dimensional neuron states of convolutional layers in memory, which is impractical for the limited size of the on-chip memory when the output tensor has a high dimension. To overcome this challenge, we propose the event-driven depth-first convolution. 

\begin{figure*}[ht]
\centering
\includegraphics[width=1.0\linewidth]{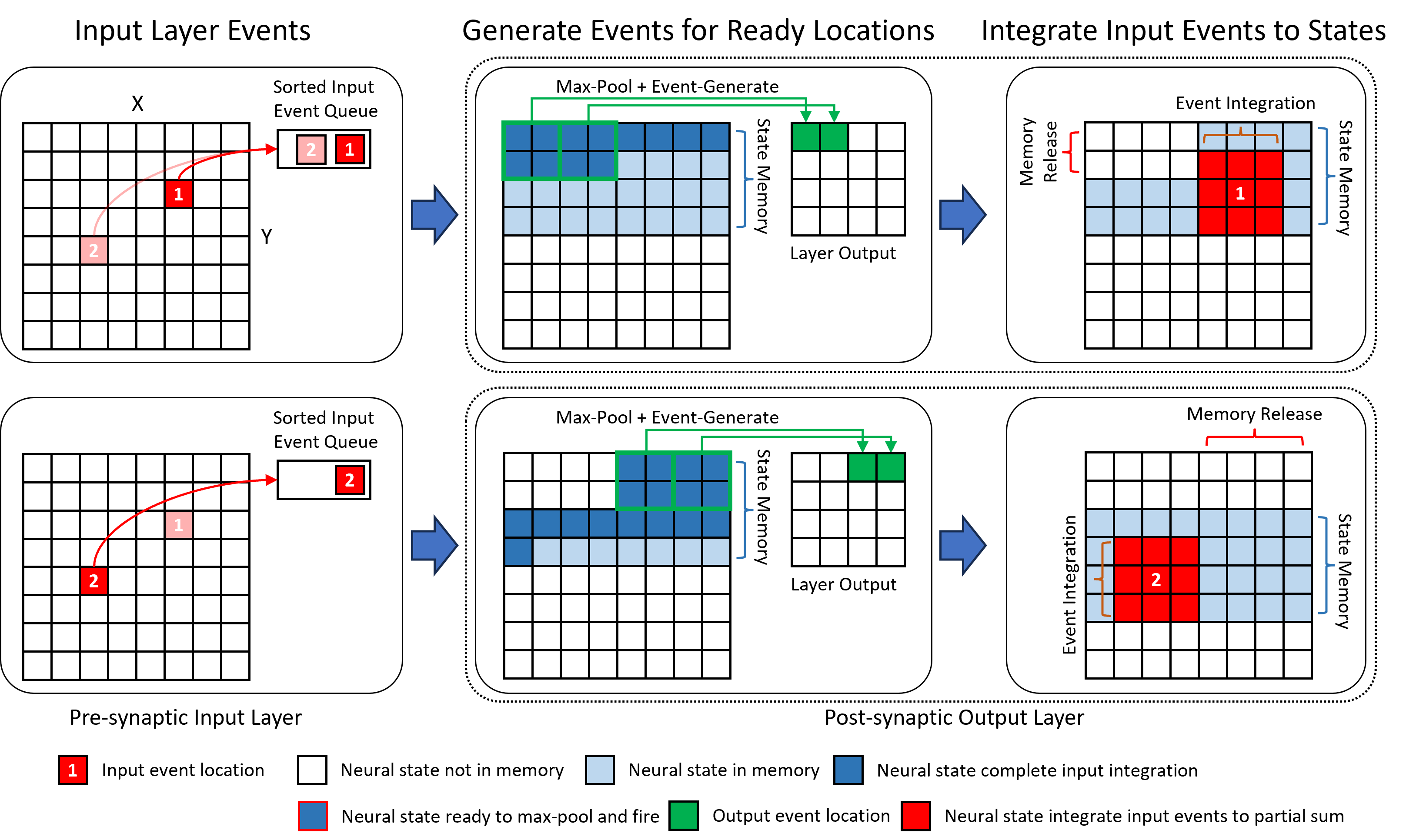}
\caption{An illustration of the event-driven depth-first convolution on SENECA. We show a fused layer combining 3$\times$3 ANN convolution and 2$\times$2 max-pooling. The layer processes input activations sequentially from the sorted input activation queue. Based on the pixel location of the new coming activation, event(s) is generated from a pixel location that has just been fully updated, and the corresponding memory spaces of the neurons at this pixel location are released. The channel dimension is omitted for simplicity.}
\label{fig:event_depth_first_conv}
\end{figure*}

Depth-first inference \cite{waeijen2021convfusion,mei2023defines} is a scheduling method in neural network inference that prioritizes the network's layer (depth) dimension by consuming \textit{ac./sp.} right after their generation. In our event-driven depth-first inference, the input \textit{ac./sp.} within a time step are assumed to be sorted in spatial order from the top-left corner of the $(X,Y)$ plane to the bottom-right corner. Under this assumption, a neuron will receive all of its input \textit{ac./sp.}s in a pre-defined order. Accordingly, its neuron state will be concluded earlier than those of spatially lower-ranked neurons (Fig. \ref{fig:event_depth_first_conv}). As a result, it can fire (conduct the event generation process) immediately after its last neuron state updating without waiting for that all the input \textit{ac./sp.} of the layer have been processed. 

After firing an ANN neuron, its neuron state is not longer required to be maintained in the memory and thus the corresponding memory block can be released. For event-driven depth-first convolution of ANN, each layer only needs to buffer a small portion of neuron states that are incomplete/partially summed (the amount of required memory increases with the kernel size). For an SNN neuron, the situation is different due to the necessity of maintaining the neuron membrane potential to the next time step. Its membrane potential is reset to zero if the membrane potential is bigger than the firing threshold, otherwise it is mutiplied with the leak parameter to get ready for the next time step. The memory for the neuron states cannot be released like ANN, \textit{i.e.} SNN needs to maintain all neuron states in the memory while ANN only need to maintain several rows of neuron (for FireNet, three rows, given the 3$\times$3 convolution kernels). This leads to the fact that, using the same memory size, ANNs can accommodate a convolutional layer whose feature map has a much bigger dimension than SNN. However, this only applied to the situation when a convolutional layer has no recurrency. For the FireNet network architecure studied in this work, the ANN has two recurrent blocks whose hidden states must be stored in the memory for the following time steps. Since we use one SENECA core to accommodate a recurrent block, the maximum resolution is constrained by the memory size of one core, the same as an SNN layer. Therefore, the maximum resolution of ANN and SNN are the same. 

Fig. \ref{fig:event_depth_first_conv} shows an example of the event-driven depth-first ANN convolutional layer with $3\times 3$ kernels and $2\times 2$ max pooling, more generic than the simple FireNet architecure (without pooling). It requires storing $(K+1)$ lines of neuron states, equal to $X\times C\times (K+1)$ neurons, where $X$ is the spatial resolution (width), $C$ is the number of channels, and $K$ is the width of the kernel. In Fig. \ref{fig:event_depth_first_conv} where $K=3$, neurons that are below the line $(X+1)$ do not need to be stored because they have not received any activation yet. Similarly, neurons that are above the line $(X-2)$ also do not need to be stored because they have already fired and do not expect to receive any more activation. 
Compared to storing all the neuron states ($X\times Y\times C$) in the on-chip memory as required by SNNs, the memory requirement for ANN neuron states is significantly reduced. For FireNet studied in this work, the ANN convolution stores $56\times 3\times 32$ neuron states while the SNN convolution stores $56\times 56\times 32$ neuron states.
This strategy enables the mapping of an ANN convolutional layer with a high spatial resolution to one SENECA core.

In event-driven depth-first convolution, the cycle of \textbf{event reception}, \textbf{neural processing}, and \textbf{event transmission} is executed as a \textit{tail-recursion} for each 2D coordinate (pixel location). Fig. \ref{fig:event_depth_first_conv} illustrates the detailed procedure of our proposed event-driven depth-first convolution on a fused convolutional (kernel size 3$\times$3, stride 1) and max-pooling (kernel size 2$\times$2, stride 2) layer. We can divide this procedure into the following phases:
\begin{itemize}

    \item When an event from the input location $(x, y)$ has been received, all the neuron states above the $(y-1)th$ row or on the left of location $(x-1, y-1)$ will not be updated further because there will not be any future incoming event that is within the kernel window view (3$\times$3 receptive field). Therefore, the event generation task will be triggered to generate the respective post-synaptic layer \textit{ac./sp.} and then free up the memory storing the neuron states (for ANN) or perform the leak operation for neuron states (for SNN). 
    
    \item After firing the fully updated neurons, the input events at location $(x, y)$ are processed. As a result, post-synaptic activity is generated at the same time as the input event trace is being processed. The event integration task integrates an input activation value to the neuron states within the 3$\times$3 spatial locations around the input location $(x, y)$. 
    
    \item If the neuron states at a spatial location have been fully updated, for ANN, the event generation task applies the activation function (\textit{e.g.}, ReLU/FATReLU) and 2$\times$2 max-pooling function to the neuron states to generate non-zero activation events. For SNN, the neuron membrane potential is compared with the firing threshold to generate binary spike events. The event transmission function packs the \textit{ac./sp.} from the same spatial neuron location (pixel location) into an event stream with shared header information of pixel coordinates and the number of \textit{ac./sp.}.
    For ANN activations, each activation is encoded into an event, 16 bits for the channel index and the rest 16 bits for its value (BFloat16). For SNN spikes, we use a bit to encode whether a channel has a spike. If there is a spike at channel $c$, then the $c$ bit will be 1, otherwise 0. For FireNet, the number of channels is 32 so an event (32 bits) can encode all channels. For a network with more channels, more than one event is required.  
    The event stream is then sent to the destination cores through the NoC.
    
\end{itemize}

\begin{figure}[!hbpt]
\begin{center}
    \centerline{\includegraphics[scale=0.44]{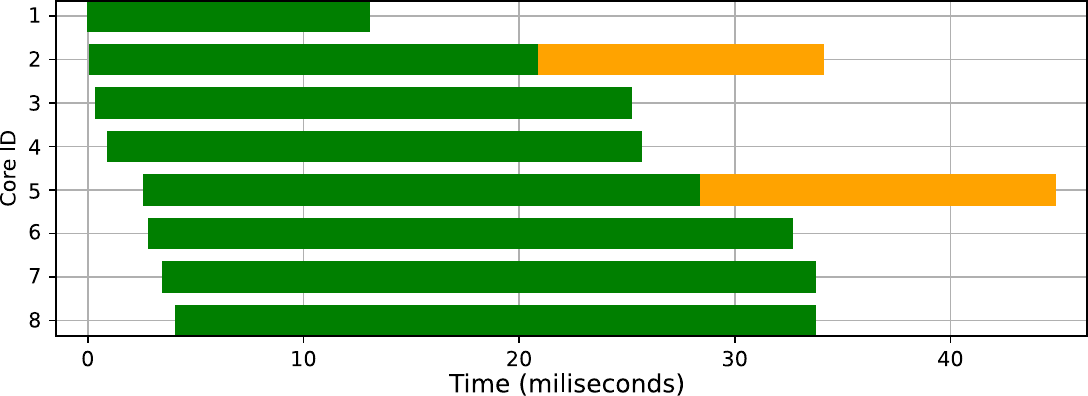}}
    \caption{Activated time of the 8 SENECA cores running the SNN FireNet processing the $56\times 56$ testing frame with $\sim$100\% average neuron density. Color green indicates the processing time of the forward convolutional layer and orange corresponds to the recurrent convolutional layer.} 
\label{fig:core_activity}
\end{center}
\end{figure}

Event-driven depth-first convolution can significantly reduce the inference latency when performing layer-to-layer event-driven data-flow processing in hardware. Traditional event-driven neuromorphic processing requires \textit{barrier} synchronization at the end of the time step before event generation and communication. This introduces an additional latency per layer that equals the time required to integrate all the input events before a neuron can fire. The lock-step processing of event reception and neural processing in depth-first convolution enables multilayered parallelism in a pipelined fashion across layers without the need for explicit per-timestep barrier synchronization primitives. As shown in Fig. \ref{fig:core_activity}, the 8 SENECA cores running the SNN FireNet are highly parallelized.

A limitation of the proposed event-driven depth-first convolution is that it requires the input \textit{ac./sp.} to be sorted and arrive in order. When using a conventional frame-based camera, this requirement is automatically satisfied. However, an additional process is needed to sort the input events of the first layer when dealing with the asynchronous measurements from an event-based camera. Nonetheless, the overhead is minimal if the input events are sparse. In this work, the overhead is not counted in the measurement of time cost because it is the same for ANN and SNN.



Despite the advantages of event-driven processing, there are significant overheads during \textbf{event reception} (unpacking/decoding each event and preparing the task) and \textbf{neural processing} (read/write neuron states per each event). The time and resources required for these steps can easily dominate the overall costs. For example, as shown in \cite{tang2023open}, a single memory access for the data movement from SRAM to registers can be more than twice the cost of an arithmetic instruction.

As mentioned, processing each \textit{ac./sp.} requires the following steps: 1) event decoding or projecting the spike address to several physical addresses of the weights and neuron states in data memory, 2) reading the relevant weights and neuron states from the data memory, 3) performing the neural calculation, and 4) writing the updated neuron states to the data memory.
To reduce the processing cost per \textit{ac./sp.}, \textit{ac./sp.} in the same time step and updating the exact same set of neurons of the next layer are combined as a group (in this work, the group size is 4). Packing such \textit{ac./sp.} that share the same destination neuron addresses significantly reduces the overhead of event decoding (step 1). Then, during the neural processing step, a neuron's state can be read once and updated multiple times by the grouped multiple \textit{ac./sp.} before it is stored back into memory, considerably reducing memory accesses (steps 2 and 4).



\section{More Details of Selected Event Frames with 56$\times$56 Pixels} \label{sec:56}

\begin{figure*}[!hbpt]
\begin{center}
    \centerline{
    \includegraphics[scale=0.58]{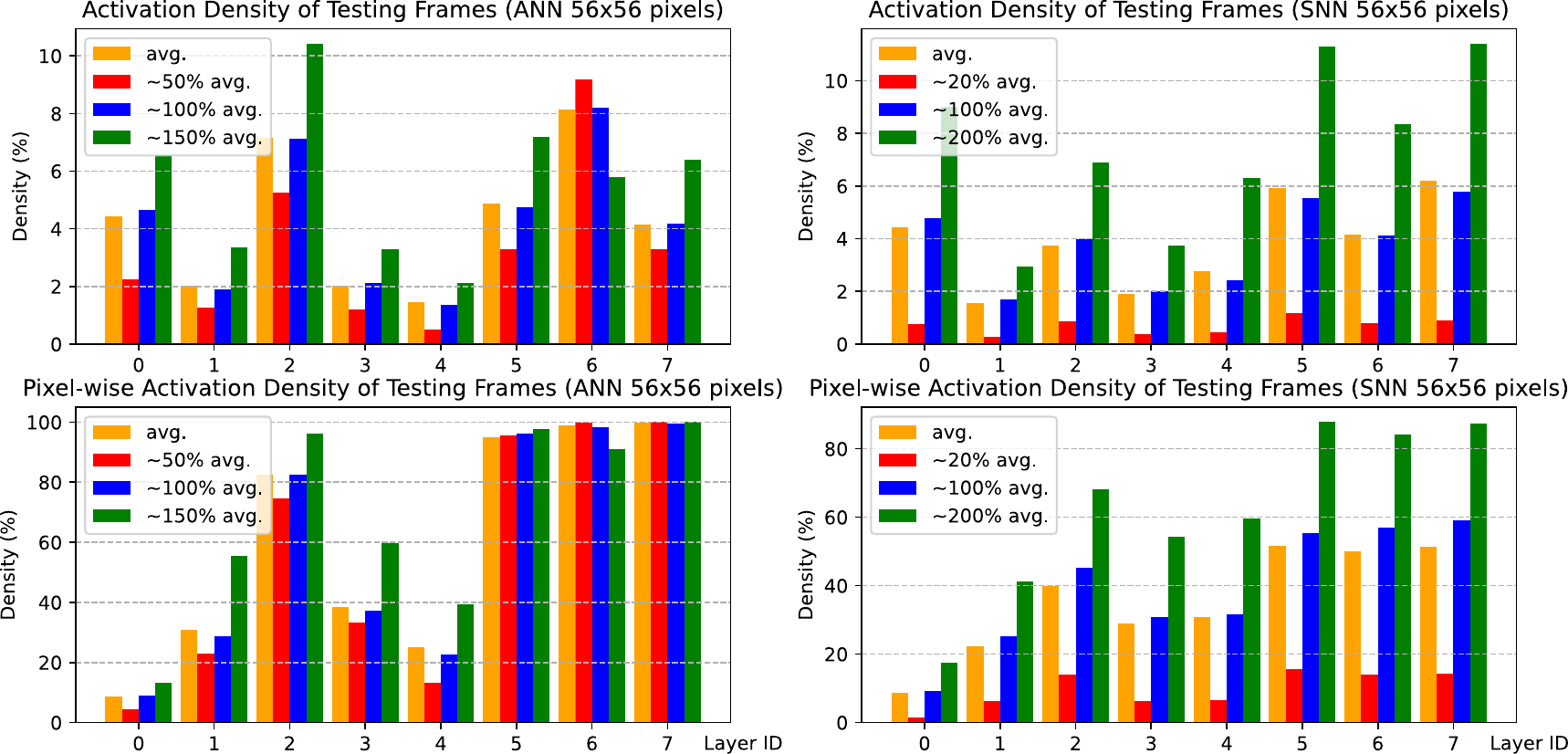}}
    \caption{Layer-wise density of the testing frames with resolution 56$\times$56 pixels. } 
\label{fig:layer_dense}
\end{center}
\end{figure*}

As introduced in Subsection \ref{subsec:network_experiments}, we select three frames with 56$\times$56 pixels for the ANN and the SNN, respectively. Here we show more information about the payer-wise \textit{ac./sp.} density produced by the selected frames in Fig. \ref{fig:layer_dense}. Comparing the ANN (two subplots on the left) and the SNN (two subplots on the right), the SNN's neuron density and pixel density are much better correlated. In contrast, the ANN's pixel density is not well correlated with neuron density, especially for the last three layers whose pixel density is almost 100\%. 
Because of the parallelization of the layers/cores shown in Fig. \ref{fig:core_activity}, those layers can be the bottleneck of the whole network inference. Therefore, as shown in Fig. \ref{fig:time_dense}, the three ANN testing frames have similar time costs.

\bibliographystyle{cas-model2-names}

\bibliography{reference}

\begin{thebibliography}{46}
\expandafter\ifx\csname natexlab\endcsname\relax\def\natexlab#1{#1}\fi
\providecommand{\url}[1]{\texttt{#1}}
\providecommand{\href}[2]{#2}
\providecommand{\path}[1]{#1}
\providecommand{\DOIprefix}{doi:}
\providecommand{\ArXivprefix}{arXiv:}
\providecommand{\URLprefix}{URL: }
\providecommand{\Pubmedprefix}{pmid:}
\providecommand{\doi}[1]{\href{http://dx.doi.org/#1}{\path{#1}}}
\providecommand{\Pubmed}[1]{\href{pmid:#1}{\path{#1}}}
\providecommand{\bibinfo}[2]{#2}
\ifx\xfnm\relax \def\xfnm[#1]{\unskip,\space#1}\fi
\bibitem[{Cao et~al.(2019)Cao, Ma, Xiao, Zhang, Liu, Zhang, Nie and Yang}]{cao2019seernet}
\bibinfo{author}{Cao, S.}, \bibinfo{author}{Ma, L.}, \bibinfo{author}{Xiao, W.}, \bibinfo{author}{Zhang, C.}, \bibinfo{author}{Liu, Y.}, \bibinfo{author}{Zhang, L.}, \bibinfo{author}{Nie, L.}, \bibinfo{author}{Yang, Z.}, \bibinfo{year}{2019}.
\newblock \bibinfo{title}{Seernet: Predicting convolutional neural network feature-map sparsity through low-bit quantization}, in: \bibinfo{booktitle}{Proceedings of the IEEE/CVF Conference on Computer Vision and Pattern Recognition}, pp. \bibinfo{pages}{11216--11225}.
\bibitem[{Chaney et~al.(2021)Chaney, Panagopoulou, Lee, Roy and Daniilidis}]{chaney2021self}
\bibinfo{author}{Chaney, K.}, \bibinfo{author}{Panagopoulou, A.}, \bibinfo{author}{Lee, C.}, \bibinfo{author}{Roy, K.}, \bibinfo{author}{Daniilidis, K.}, \bibinfo{year}{2021}.
\newblock \bibinfo{title}{Self-supervised optical flow with spiking neural networks and event based cameras}, in: \bibinfo{booktitle}{2021 IEEE/RSJ International Conference on Intelligent Robots and Systems (IROS)}, \bibinfo{organization}{IEEE}. pp. \bibinfo{pages}{5892--5899}.
\bibitem[{Chen et~al.(2019)Chen, Yang, Emer and Sze}]{chen2019eyeriss}
\bibinfo{author}{Chen, Y.H.}, \bibinfo{author}{Yang, T.J.}, \bibinfo{author}{Emer, J.}, \bibinfo{author}{Sze, V.}, \bibinfo{year}{2019}.
\newblock \bibinfo{title}{Eyeriss v2: A flexible accelerator for emerging deep neural networks on mobile devices}.
\newblock \bibinfo{journal}{IEEE Journal on Emerging and Selected Topics in Circuits and Systems} \bibinfo{volume}{9}, \bibinfo{pages}{292--308}.
\bibitem[{Cuadrado et~al.(2023)Cuadrado, Ran{\c{c}}on, Cottereau, Barranco and Masquelier}]{cuadrado2023optical}
\bibinfo{author}{Cuadrado, J.}, \bibinfo{author}{Ran{\c{c}}on, U.}, \bibinfo{author}{Cottereau, B.R.}, \bibinfo{author}{Barranco, F.}, \bibinfo{author}{Masquelier, T.}, \bibinfo{year}{2023}.
\newblock \bibinfo{title}{Optical flow estimation from event-based cameras and spiking neural networks}.
\newblock \bibinfo{journal}{Frontiers in Neuroscience} \bibinfo{volume}{17}, \bibinfo{pages}{1160034}.
\bibitem[{Dampfhoffer et~al.(2022)Dampfhoffer, Mesquida, Valentian and Anghel}]{dampfhoffer2022snns}
\bibinfo{author}{Dampfhoffer, M.}, \bibinfo{author}{Mesquida, T.}, \bibinfo{author}{Valentian, A.}, \bibinfo{author}{Anghel, L.}, \bibinfo{year}{2022}.
\newblock \bibinfo{title}{Are snns really more energy-efficient than anns? an in-depth hardware-aware study}.
\newblock \bibinfo{journal}{IEEE Transactions on Emerging Topics in Computational Intelligence} .
\bibitem[{Davidson and Furber(2021)}]{davidson2021comparison}
\bibinfo{author}{Davidson, S.}, \bibinfo{author}{Furber, S.B.}, \bibinfo{year}{2021}.
\newblock \bibinfo{title}{Comparison of artificial and spiking neural networks on digital hardware}.
\newblock \bibinfo{journal}{Frontiers in Neuroscience} \bibinfo{volume}{15}, \bibinfo{pages}{651141}.
\bibitem[{Davies et~al.(2021)Davies, Wild, Orchard, Sandamirskaya, Guerra, Joshi, Plank and Risbud}]{davies2021advancing}
\bibinfo{author}{Davies, M.}, \bibinfo{author}{Wild, A.}, \bibinfo{author}{Orchard, G.}, \bibinfo{author}{Sandamirskaya, Y.}, \bibinfo{author}{Guerra, G.A.F.}, \bibinfo{author}{Joshi, P.}, \bibinfo{author}{Plank, P.}, \bibinfo{author}{Risbud, S.R.}, \bibinfo{year}{2021}.
\newblock \bibinfo{title}{Advancing neuromorphic computing with loihi: A survey of results and outlook}.
\newblock \bibinfo{journal}{Proceedings of the IEEE} \bibinfo{volume}{109}, \bibinfo{pages}{911--934}.
\bibitem[{Delmerico et~al.(2019)Delmerico, Cieslewski, Rebecq, Faessler and Scaramuzza}]{delmerico2019we}
\bibinfo{author}{Delmerico, J.}, \bibinfo{author}{Cieslewski, T.}, \bibinfo{author}{Rebecq, H.}, \bibinfo{author}{Faessler, M.}, \bibinfo{author}{Scaramuzza, D.}, \bibinfo{year}{2019}.
\newblock \bibinfo{title}{Are we ready for autonomous drone racing? the uzh-fpv drone racing dataset}, in: \bibinfo{booktitle}{2019 International Conference on Robotics and Automation (ICRA)}, \bibinfo{organization}{IEEE}. pp. \bibinfo{pages}{6713--6719}.
\bibitem[{Deng et~al.(2020)Deng, Wu, Hu, Liang, Ding, Li, Zhao, Li and Xie}]{deng2020rethinking}
\bibinfo{author}{Deng, L.}, \bibinfo{author}{Wu, Y.}, \bibinfo{author}{Hu, X.}, \bibinfo{author}{Liang, L.}, \bibinfo{author}{Ding, Y.}, \bibinfo{author}{Li, G.}, \bibinfo{author}{Zhao, G.}, \bibinfo{author}{Li, P.}, \bibinfo{author}{Xie, Y.}, \bibinfo{year}{2020}.
\newblock \bibinfo{title}{Rethinking the performance comparison between snns and anns}.
\newblock \bibinfo{journal}{Neural networks} \bibinfo{volume}{121}, \bibinfo{pages}{294--307}.
\bibitem[{Furber and Bogdan(2020)}]{furber2020spinnaker}
\bibinfo{author}{Furber, S.}, \bibinfo{author}{Bogdan, P.}, \bibinfo{year}{2020}.
\newblock \bibinfo{title}{Spinnaker-a spiking neural network architecture}.
\newblock \bibinfo{publisher}{Now publishers}.
\bibitem[{Gallego et~al.(2019)Gallego, Gehrig and Scaramuzza}]{gallego2019focus}
\bibinfo{author}{Gallego, G.}, \bibinfo{author}{Gehrig, M.}, \bibinfo{author}{Scaramuzza, D.}, \bibinfo{year}{2019}.
\newblock \bibinfo{title}{Focus is all you need: Loss functions for event-based vision}, in: \bibinfo{booktitle}{Proceedings of the IEEE/CVF Conference on Computer Vision and Pattern Recognition}, pp. \bibinfo{pages}{12280--12289}.
\bibitem[{Gehrig et~al.(2021)Gehrig, Millh{\"a}usler, Gehrig and Scaramuzza}]{gehrig2021raft}
\bibinfo{author}{Gehrig, M.}, \bibinfo{author}{Millh{\"a}usler, M.}, \bibinfo{author}{Gehrig, D.}, \bibinfo{author}{Scaramuzza, D.}, \bibinfo{year}{2021}.
\newblock \bibinfo{title}{E-raft: Dense optical flow from event cameras}, in: \bibinfo{booktitle}{2021 International Conference on 3D Vision (3DV)}, \bibinfo{organization}{IEEE}. pp. \bibinfo{pages}{197--206}.
\bibitem[{Georgiadis(2019)}]{georgiadis2019accelerating}
\bibinfo{author}{Georgiadis, G.}, \bibinfo{year}{2019}.
\newblock \bibinfo{title}{Accelerating convolutional neural networks via activation map compression}, in: \bibinfo{booktitle}{Proceedings of the IEEE/CVF Conference on Computer Vision and Pattern Recognition}, pp. \bibinfo{pages}{7085--7095}.
\bibitem[{Grimaldi et~al.(2023)Grimaldi, Ganji, Lazarevich and Sah}]{grimaldi2023accelerating}
\bibinfo{author}{Grimaldi, M.}, \bibinfo{author}{Ganji, D.C.}, \bibinfo{author}{Lazarevich, I.}, \bibinfo{author}{Sah, S.}, \bibinfo{year}{2023}.
\newblock \bibinfo{title}{Accelerating deep neural networks via semi-structured activation sparsity}, in: \bibinfo{booktitle}{Proceedings of the IEEE/CVF International Conference on Computer Vision}, pp. \bibinfo{pages}{1179--1188}.
\bibitem[{Hagenaars et~al.(2021)Hagenaars, Paredes-Vall{\'e}s and De~Croon}]{hagenaars2021self}
\bibinfo{author}{Hagenaars, J.}, \bibinfo{author}{Paredes-Vall{\'e}s, F.}, \bibinfo{author}{De~Croon, G.}, \bibinfo{year}{2021}.
\newblock \bibinfo{title}{Self-supervised learning of event-based optical flow with spiking neural networks}.
\newblock \bibinfo{journal}{Advances in Neural Information Processing Systems} \bibinfo{volume}{34}, \bibinfo{pages}{7167--7179}.
\bibitem[{Hoefler et~al.(2021)Hoefler, Alistarh, Ben-Nun, Dryden and Peste}]{hoefler2021sparsity}
\bibinfo{author}{Hoefler, T.}, \bibinfo{author}{Alistarh, D.}, \bibinfo{author}{Ben-Nun, T.}, \bibinfo{author}{Dryden, N.}, \bibinfo{author}{Peste, A.}, \bibinfo{year}{2021}.
\newblock \bibinfo{title}{Sparsity in deep learning: Pruning and growth for efficient inference and training in neural networks}.
\newblock \bibinfo{journal}{Journal of Machine Learning Research} \bibinfo{volume}{22}, \bibinfo{pages}{1--124}.
\bibitem[{Horowitz(2014)}]{horowitz20141}
\bibinfo{author}{Horowitz, M.}, \bibinfo{year}{2014}.
\newblock \bibinfo{title}{1.1 computing's energy problem (and what we can do about it)}, in: \bibinfo{booktitle}{2014 IEEE international solid-state circuits conference digest of technical papers (ISSCC)}, \bibinfo{organization}{IEEE}. pp. \bibinfo{pages}{10--14}.
\bibitem[{Hoyer(2004)}]{hoyer2004non}
\bibinfo{author}{Hoyer, P.O.}, \bibinfo{year}{2004}.
\newblock \bibinfo{title}{Non-negative matrix factorization with sparseness constraints.}
\newblock \bibinfo{journal}{Journal of machine learning research} \bibinfo{volume}{5}.
\bibitem[{Kosta and Roy(2023)}]{kosta2023adaptive}
\bibinfo{author}{Kosta, A.K.}, \bibinfo{author}{Roy, K.}, \bibinfo{year}{2023}.
\newblock \bibinfo{title}{Adaptive-spikenet: event-based optical flow estimation using spiking neural networks with learnable neuronal dynamics}, in: \bibinfo{booktitle}{2023 IEEE International Conference on Robotics and Automation (ICRA)}, \bibinfo{organization}{IEEE}. pp. \bibinfo{pages}{6021--6027}.
\bibitem[{Kurtz et~al.(2020)Kurtz, Kopinsky, Gelashvili, Matveev, Carr, Goin, Leiserson, Moore, Shavit and Alistarh}]{kurtz2020inducing}
\bibinfo{author}{Kurtz, M.}, \bibinfo{author}{Kopinsky, J.}, \bibinfo{author}{Gelashvili, R.}, \bibinfo{author}{Matveev, A.}, \bibinfo{author}{Carr, J.}, \bibinfo{author}{Goin, M.}, \bibinfo{author}{Leiserson, W.}, \bibinfo{author}{Moore, S.}, \bibinfo{author}{Shavit, N.}, \bibinfo{author}{Alistarh, D.}, \bibinfo{year}{2020}.
\newblock \bibinfo{title}{Inducing and exploiting activation sparsity for fast inference on deep neural networks}, in: \bibinfo{booktitle}{International Conference on Machine Learning}, \bibinfo{organization}{PMLR}. pp. \bibinfo{pages}{5533--5543}.
\bibitem[{Lee et~al.(2020)Lee, Kosta, Zhu, Chaney, Daniilidis and Roy}]{lee2020spike}
\bibinfo{author}{Lee, C.}, \bibinfo{author}{Kosta, A.K.}, \bibinfo{author}{Zhu, A.Z.}, \bibinfo{author}{Chaney, K.}, \bibinfo{author}{Daniilidis, K.}, \bibinfo{author}{Roy, K.}, \bibinfo{year}{2020}.
\newblock \bibinfo{title}{Spike-flownet: event-based optical flow estimation with energy-efficient hybrid neural networks}, in: \bibinfo{booktitle}{European Conference on Computer Vision}, \bibinfo{organization}{Springer}. pp. \bibinfo{pages}{366--382}.
\bibitem[{Lee et~al.(2021)Lee, Kim, Lee, Baek and Kim}]{lee2021accurate}
\bibinfo{author}{Lee, H.}, \bibinfo{author}{Kim, C.}, \bibinfo{author}{Lee, S.}, \bibinfo{author}{Baek, E.}, \bibinfo{author}{Kim, J.}, \bibinfo{year}{2021}.
\newblock \bibinfo{title}{An accurate and fair evaluation methodology for snn-based inferencing with full-stack hardware design space explorations}.
\newblock \bibinfo{journal}{Neurocomputing} \bibinfo{volume}{455}, \bibinfo{pages}{125--138}.
\bibitem[{Lemaire et~al.(2022)Lemaire, Miramond, Bilavarn, Saoud and Abderrahmane}]{lemaire2022synaptic}
\bibinfo{author}{Lemaire, E.}, \bibinfo{author}{Miramond, B.}, \bibinfo{author}{Bilavarn, S.}, \bibinfo{author}{Saoud, H.}, \bibinfo{author}{Abderrahmane, N.}, \bibinfo{year}{2022}.
\newblock \bibinfo{title}{Synaptic activity and hardware footprint of spiking neural networks in digital neuromorphic systems}.
\newblock \bibinfo{journal}{ACM Transactions on Embedded Computing Systems} \bibinfo{volume}{21}, \bibinfo{pages}{1--26}.
\bibitem[{Li et~al.(2023)Li, You, Bhojanapalli, Li, Rawat, Reddi, Ye, Chern, Yu, Guo et~al.}]{li2023lazy}
\bibinfo{author}{Li, Z.}, \bibinfo{author}{You, C.}, \bibinfo{author}{Bhojanapalli, S.}, \bibinfo{author}{Li, D.}, \bibinfo{author}{Rawat, A.S.}, \bibinfo{author}{Reddi, S.J.}, \bibinfo{author}{Ye, K.}, \bibinfo{author}{Chern, F.}, \bibinfo{author}{Yu, F.}, \bibinfo{author}{Guo, R.}, et~al., \bibinfo{year}{2023}.
\newblock \bibinfo{title}{The lazy neuron phenomenon: On emergence of activation sparsity in transformers}, in: \bibinfo{booktitle}{The Eleventh International Conference on Learning Representations}.
\bibitem[{Mei et~al.(2023)Mei, Goetschalckx, Symons and Verhelst}]{mei2023defines}
\bibinfo{author}{Mei, L.}, \bibinfo{author}{Goetschalckx, K.}, \bibinfo{author}{Symons, A.}, \bibinfo{author}{Verhelst, M.}, \bibinfo{year}{2023}.
\newblock \bibinfo{title}{{DeFiNES}: Enabling fast exploration of the depth-first scheduling space for dnn accelerators through analytical modeling}, in: \bibinfo{booktitle}{2023 IEEE International Symposium on High-Performance Computer Architecture (HPCA)}, \bibinfo{organization}{IEEE}. pp. \bibinfo{pages}{570--583}.
\bibitem[{Modha et~al.(2023)Modha, Akopyan, Andreopoulos, Appuswamy, Arthur, Cassidy, Datta, DeBole, Esser, Otero et~al.}]{modha2023neural}
\bibinfo{author}{Modha, D.S.}, \bibinfo{author}{Akopyan, F.}, \bibinfo{author}{Andreopoulos, A.}, \bibinfo{author}{Appuswamy, R.}, \bibinfo{author}{Arthur, J.V.}, \bibinfo{author}{Cassidy, A.S.}, \bibinfo{author}{Datta, P.}, \bibinfo{author}{DeBole, M.V.}, \bibinfo{author}{Esser, S.K.}, \bibinfo{author}{Otero, C.O.}, et~al., \bibinfo{year}{2023}.
\newblock \bibinfo{title}{Neural inference at the frontier of energy, space, and time}.
\newblock \bibinfo{journal}{Science} \bibinfo{volume}{382}, \bibinfo{pages}{329--335}.
\bibitem[{Parashar et~al.(2017)Parashar, Rhu, Mukkara, Puglielli, Venkatesan, Khailany, Emer, Keckler and Dally}]{parashar2017scnn}
\bibinfo{author}{Parashar, A.}, \bibinfo{author}{Rhu, M.}, \bibinfo{author}{Mukkara, A.}, \bibinfo{author}{Puglielli, A.}, \bibinfo{author}{Venkatesan, R.}, \bibinfo{author}{Khailany, B.}, \bibinfo{author}{Emer, J.}, \bibinfo{author}{Keckler, S.W.}, \bibinfo{author}{Dally, W.J.}, \bibinfo{year}{2017}.
\newblock \bibinfo{title}{Scnn: An accelerator for compressed-sparse convolutional neural networks}.
\newblock \bibinfo{journal}{ACM SIGARCH computer architecture news} \bibinfo{volume}{45}, \bibinfo{pages}{27--40}.
\bibitem[{Paredes-Vall{\'e}s et~al.(2023)Paredes-Vall{\'e}s, Hagenaars, Dupeyroux, Stroobants, Xu and de~Croon}]{paredes2023fully}
\bibinfo{author}{Paredes-Vall{\'e}s, F.}, \bibinfo{author}{Hagenaars, J.}, \bibinfo{author}{Dupeyroux, J.}, \bibinfo{author}{Stroobants, S.}, \bibinfo{author}{Xu, Y.}, \bibinfo{author}{de~Croon, G.}, \bibinfo{year}{2023}.
\newblock \bibinfo{title}{Fully neuromorphic vision and control for autonomous drone flight}.
\newblock \bibinfo{journal}{arXiv preprint arXiv:2303.08778} .
\bibitem[{Paredes-Vall\'es et~al.(2023)Paredes-Vall\'es, Scheper, De~Wagter and de~Croon}]{Paredes-Valles_2023_ICCV}
\bibinfo{author}{Paredes-Vall\'es, F.}, \bibinfo{author}{Scheper, K.Y.W.}, \bibinfo{author}{De~Wagter, C.}, \bibinfo{author}{de~Croon, G.C.H.E.}, \bibinfo{year}{2023}.
\newblock \bibinfo{title}{Taming contrast maximization for learning sequential, low-latency, event-based optical flow}, in: \bibinfo{booktitle}{Proceedings of the IEEE/CVF International Conference on Computer Vision (ICCV)}, pp. \bibinfo{pages}{9695--9705}.
\bibitem[{Ponghiran et~al.(2023)Ponghiran, Liyanagedera and Roy}]{ponghiran2023event}
\bibinfo{author}{Ponghiran, W.}, \bibinfo{author}{Liyanagedera, C.M.}, \bibinfo{author}{Roy, K.}, \bibinfo{year}{2023}.
\newblock \bibinfo{title}{Event-based temporally dense optical flow estimation with sequential learning}, in: \bibinfo{booktitle}{Proceedings of the IEEE/CVF International Conference on Computer Vision}, pp. \bibinfo{pages}{9827--9836}.
\bibitem[{Rhodes et~al.(2018)Rhodes, Bogdan, Brenninkmeijer, Davidson, Fellows, Gait, Lester, Mikaitis, Plana, Rowley et~al.}]{rhodes2018spynnaker}
\bibinfo{author}{Rhodes, O.}, \bibinfo{author}{Bogdan, P.A.}, \bibinfo{author}{Brenninkmeijer, C.}, \bibinfo{author}{Davidson, S.}, \bibinfo{author}{Fellows, D.}, \bibinfo{author}{Gait, A.}, \bibinfo{author}{Lester, D.R.}, \bibinfo{author}{Mikaitis, M.}, \bibinfo{author}{Plana, L.A.}, \bibinfo{author}{Rowley, A.G.}, et~al., \bibinfo{year}{2018}.
\newblock \bibinfo{title}{spynnaker: a software package for running pynn simulations on spinnaker}.
\newblock \bibinfo{journal}{Frontiers in neuroscience} \bibinfo{volume}{12}, \bibinfo{pages}{816}.
\bibitem[{Runwal et~al.(2023)Runwal, Pedapati and Chen}]{runwal2023parameter}
\bibinfo{author}{Runwal, B.}, \bibinfo{author}{Pedapati, T.}, \bibinfo{author}{Chen, P.Y.}, \bibinfo{year}{2023}.
\newblock \bibinfo{title}{Parameter efficient finetuning for reducing activation density in transformers}, in: \bibinfo{booktitle}{Annual Conference on Neural Information Processing Systems}.
\bibitem[{Schnider et~al.(2023)Schnider, Wo{\'z}niak, Gehrig, Lecomte, Von~Arnim, Benini, Scaramuzza and Pantazi}]{schnider2023neuromorphic}
\bibinfo{author}{Schnider, Y.}, \bibinfo{author}{Wo{\'z}niak, S.}, \bibinfo{author}{Gehrig, M.}, \bibinfo{author}{Lecomte, J.}, \bibinfo{author}{Von~Arnim, A.}, \bibinfo{author}{Benini, L.}, \bibinfo{author}{Scaramuzza, D.}, \bibinfo{author}{Pantazi, A.}, \bibinfo{year}{2023}.
\newblock \bibinfo{title}{Neuromorphic optical flow and real-time implementation with event cameras}, in: \bibinfo{booktitle}{Proceedings of the IEEE/CVF Conference on Computer Vision and Pattern Recognition}, pp. \bibinfo{pages}{4128--4137}.
\bibitem[{Sengupta et~al.(2019)Sengupta, Ye, Wang, Liu and Roy}]{sengupta2019going}
\bibinfo{author}{Sengupta, A.}, \bibinfo{author}{Ye, Y.}, \bibinfo{author}{Wang, R.}, \bibinfo{author}{Liu, C.}, \bibinfo{author}{Roy, K.}, \bibinfo{year}{2019}.
\newblock \bibinfo{title}{Going deeper in spiking neural networks: Vgg and residual architectures}.
\newblock \bibinfo{journal}{Frontiers in neuroscience} \bibinfo{volume}{13}, \bibinfo{pages}{95}.
\bibitem[{Shiba et~al.(2022)Shiba, Aoki and Gallego}]{shiba2022secrets}
\bibinfo{author}{Shiba, S.}, \bibinfo{author}{Aoki, Y.}, \bibinfo{author}{Gallego, G.}, \bibinfo{year}{2022}.
\newblock \bibinfo{title}{Secrets of event-based optical flow}, in: \bibinfo{booktitle}{European Conference on Computer Vision}, \bibinfo{organization}{Springer}. pp. \bibinfo{pages}{628--645}.
\bibitem[{Tang et~al.(2023a)Tang, Safa, Shidqi, Detterer, Traferro, Konijnenburg, Sifalakis, van Schaik and Yousefzadeh}]{tang2023open}
\bibinfo{author}{Tang, G.}, \bibinfo{author}{Safa, A.}, \bibinfo{author}{Shidqi, K.}, \bibinfo{author}{Detterer, P.}, \bibinfo{author}{Traferro, S.}, \bibinfo{author}{Konijnenburg, M.}, \bibinfo{author}{Sifalakis, M.}, \bibinfo{author}{van Schaik, G.J.}, \bibinfo{author}{Yousefzadeh, A.}, \bibinfo{year}{2023}a.
\newblock \bibinfo{title}{Open the box of digital neuromorphic processor: Towards effective algorithm-hardware co-design}, in: \bibinfo{booktitle}{2023 IEEE International Symposium on Circuits and Systems (ISCAS)}, \bibinfo{organization}{IEEE}. pp. \bibinfo{pages}{1--5}.
\bibitem[{Tang et~al.(2023b)Tang, Vadivel, Xu, Bilgic, Shidqi, Detterer, Traferro, Konijnenburg, Sifalakis, van Schaik et~al.}]{tang2023seneca}
\bibinfo{author}{Tang, G.}, \bibinfo{author}{Vadivel, K.}, \bibinfo{author}{Xu, Y.}, \bibinfo{author}{Bilgic, R.}, \bibinfo{author}{Shidqi, K.}, \bibinfo{author}{Detterer, P.}, \bibinfo{author}{Traferro, S.}, \bibinfo{author}{Konijnenburg, M.}, \bibinfo{author}{Sifalakis, M.}, \bibinfo{author}{van Schaik, G.J.}, et~al., \bibinfo{year}{2023}b.
\newblock \bibinfo{title}{Seneca: building a fully digital neuromorphic processor, design trade-offs and challenges}.
\newblock \bibinfo{journal}{Frontiers in Neuroscience} \bibinfo{volume}{17}.
\bibitem[{Waeijen et~al.(2021)Waeijen, Sioutas, Peemen, Lindwer and Corporaal}]{waeijen2021convfusion}
\bibinfo{author}{Waeijen, L.}, \bibinfo{author}{Sioutas, S.}, \bibinfo{author}{Peemen, M.}, \bibinfo{author}{Lindwer, M.}, \bibinfo{author}{Corporaal, H.}, \bibinfo{year}{2021}.
\newblock \bibinfo{title}{{ConvFusion}: A model for layer fusion in convolutional neural networks}.
\newblock \bibinfo{journal}{IEEE Access} \bibinfo{volume}{9}, \bibinfo{pages}{168245--168267}.
\bibitem[{Wu et~al.(2019)Wu, Deng, Li, Zhu, Xie and Shi}]{wu2019direct}
\bibinfo{author}{Wu, Y.}, \bibinfo{author}{Deng, L.}, \bibinfo{author}{Li, G.}, \bibinfo{author}{Zhu, J.}, \bibinfo{author}{Xie, Y.}, \bibinfo{author}{Shi, L.}, \bibinfo{year}{2019}.
\newblock \bibinfo{title}{Direct training for spiking neural networks: Faster, larger, better}, in: \bibinfo{booktitle}{Proceedings of the AAAI conference on artificial intelligence}, pp. \bibinfo{pages}{1311--1318}.
\bibitem[{Xu et~al.(2024)Xu, Shidqi, van Schaik, Bilgic, Dobrita, Wang, Meijer, Nembhani, Arjmand, Martinello et~al.}]{xu2024optimizing}
\bibinfo{author}{Xu, Y.}, \bibinfo{author}{Shidqi, K.}, \bibinfo{author}{van Schaik, G.J.}, \bibinfo{author}{Bilgic, R.}, \bibinfo{author}{Dobrita, A.}, \bibinfo{author}{Wang, S.}, \bibinfo{author}{Meijer, R.}, \bibinfo{author}{Nembhani, P.}, \bibinfo{author}{Arjmand, C.}, \bibinfo{author}{Martinello, P.}, et~al., \bibinfo{year}{2024}.
\newblock \bibinfo{title}{Optimizing event-based neural networks on digital neuromorphic architecture: a comprehensive design space exploration}.
\newblock \bibinfo{journal}{Frontiers in Neuroscience} \bibinfo{volume}{18}, \bibinfo{pages}{1335422}.
\bibitem[{Yang et~al.(2021)Yang, Mao, Wang and Hai}]{yang2021dynamic}
\bibinfo{author}{Yang, Q.}, \bibinfo{author}{Mao, J.}, \bibinfo{author}{Wang, Z.}, \bibinfo{author}{Hai, H.L.}, \bibinfo{year}{2021}.
\newblock \bibinfo{title}{Dynamic regularization on activation sparsity for neural network efficiency improvement}.
\newblock \bibinfo{journal}{ACM Journal on Emerging Technologies in Computing Systems (JETC)} \bibinfo{volume}{17}, \bibinfo{pages}{1--16}.
\bibitem[{Yousefzadeh et~al.(2017)Yousefzadeh, Jab{\l}o{\'n}ski, Iakymchuk, Linares-Barranco, Rosado, Plana, Temple, Serrano-Gotarredona, Furber and Linares-Barranco}]{AER2017}
\bibinfo{author}{Yousefzadeh, A.}, \bibinfo{author}{Jab{\l}o{\'n}ski, M.}, \bibinfo{author}{Iakymchuk, T.}, \bibinfo{author}{Linares-Barranco, A.}, \bibinfo{author}{Rosado, A.}, \bibinfo{author}{Plana, L.A.}, \bibinfo{author}{Temple, S.}, \bibinfo{author}{Serrano-Gotarredona, T.}, \bibinfo{author}{Furber, S.B.}, \bibinfo{author}{Linares-Barranco, B.}, \bibinfo{year}{2017}.
\newblock \bibinfo{title}{On multiple aer handshaking channels over high-speed bit-serial bidirectional lvds links with flow-control and clock-correction on commercial fpgas for scalable neuromorphic systems}.
\newblock \bibinfo{journal}{IEEE transactions on biomedical circuits and systems} \bibinfo{volume}{11}, \bibinfo{pages}{1133--1147}.
\bibitem[{Yousefzadeh et~al.(2022)Yousefzadeh, Van~Schaik, Tahghighi, Detterer, Traferro, Hijdra, Stuijt, Corradi, Sifalakis and Konijnenburg}]{yousefzadeh2022seneca}
\bibinfo{author}{Yousefzadeh, A.}, \bibinfo{author}{Van~Schaik, G.J.}, \bibinfo{author}{Tahghighi, M.}, \bibinfo{author}{Detterer, P.}, \bibinfo{author}{Traferro, S.}, \bibinfo{author}{Hijdra, M.}, \bibinfo{author}{Stuijt, J.}, \bibinfo{author}{Corradi, F.}, \bibinfo{author}{Sifalakis, M.}, \bibinfo{author}{Konijnenburg, M.}, \bibinfo{year}{2022}.
\newblock \bibinfo{title}{Seneca: Scalable energy-efficient neuromorphic computer architecture}, in: \bibinfo{booktitle}{2022 IEEE 4th International Conference on Artificial Intelligence Circuits and Systems (AICAS)}, \bibinfo{organization}{IEEE}. pp. \bibinfo{pages}{371--374}.
\bibitem[{Zhu et~al.(2018)Zhu, Thakur, {\"O}zaslan, Pfrommer, Kumar and Daniilidis}]{zhu2018multivehicle}
\bibinfo{author}{Zhu, A.Z.}, \bibinfo{author}{Thakur, D.}, \bibinfo{author}{{\"O}zaslan, T.}, \bibinfo{author}{Pfrommer, B.}, \bibinfo{author}{Kumar, V.}, \bibinfo{author}{Daniilidis, K.}, \bibinfo{year}{2018}.
\newblock \bibinfo{title}{The multivehicle stereo event camera dataset: An event camera dataset for 3d perception}.
\newblock \bibinfo{journal}{IEEE Robotics and Automation Letters} \bibinfo{volume}{3}, \bibinfo{pages}{2032--2039}.
\bibitem[{Zhu and Yuan(2018)}]{zhu2018ev}
\bibinfo{author}{Zhu, A.Z.}, \bibinfo{author}{Yuan, L.}, \bibinfo{year}{2018}.
\newblock \bibinfo{title}{Ev-flownet: Self-supervised optical flow estimation for event-based cameras}, in: \bibinfo{booktitle}{Robotics: Science and Systems}.
\bibitem[{Zhu et~al.(2019)Zhu, Yuan, Chaney and Daniilidis}]{zhu2019unsupervised}
\bibinfo{author}{Zhu, A.Z.}, \bibinfo{author}{Yuan, L.}, \bibinfo{author}{Chaney, K.}, \bibinfo{author}{Daniilidis, K.}, \bibinfo{year}{2019}.
\newblock \bibinfo{title}{Unsupervised event-based learning of optical flow, depth, and egomotion}, in: \bibinfo{booktitle}{Proceedings of the IEEE/CVF Conference on Computer Vision and Pattern Recognition}, pp. \bibinfo{pages}{989--997}.

\end{thebibliography}





\end{document}